\documentclass[10pt,twocolumn,letterpaper]{article}

\usepackage{cvpr}
\usepackage{times}
\usepackage{epsfig}
\usepackage{graphicx}
\usepackage{amsmath}
\usepackage{amssymb}
\usepackage{subfig}
\usepackage{epstopdf}
\usepackage{setspace}
\usepackage{float}
\usepackage{stfloats}
\usepackage{algorithm}
\usepackage{algorithmic}
\pdfoutput=1 

\usepackage[breaklinks=true,bookmarks=false,,colorlinks,linkcolor=red]{hyperref}

\cvprfinalcopy 

\begin{document}

\title{Analyzing Green View Index and Green View Index best path using Google Street View and deep learning}



\author{Jiahao Zhang\thanks{Corresponding author: zjhambition@gmail.com}\\
Osaka University\\
{\tt\small zjhambition@gmail.com}
\and
Anqi Hu\\
Osaka University\\
{\tt\small anqihu1028@gmail.com}
}

\maketitle

\begin{abstract}
As an important part of urban landscape research, analyzing and studying street-level greenery can increase the understanding of a city's greenery, contributing to better urban living environment planning and design. Planning the best path of urban greenery is a means to effectively maximize the use of urban greenery, which plays a positive role in the physical and mental health of urban residents and the path planning of visitors. In this paper, we used Google Street View (GSV) to obtain street view images of Osaka City. The semantic segmentation model is adopted to segment the street view images and analyze the Green View Index (GVI) of Osaka City. Based on the GVI, we take advantage of the adjacency matrix and Floyd-Warshall Algorithm to calculate Green View Index best path, solving the limitations of ArcGIS software. Our analysis not only allows the calculation of specific routes for the GVI best paths but also realizes the visualization and integration of neighborhood urban greenery. By summarizing all the data, we can conduct an intuitive feeling and objective analysis of the street-level greenery in the research area. Based on this, such as urban residents and visitors can maximize the available natural resources for a better life. The dataset and code are available at \url{https://github.com/Jackieam/GVI-Best-Path}. 
\end{abstract}


\setcounter{section}{0}
\section{Introduction}
    The street-level urban greenery is a necessary part of the urban landscape \cite{li2015assessing}. The environmental resources that people live in affect their lifestyle. Several studies have shown a positive relationship between the availability of street-level urban greenery and the health of urban residents, which may provide opportunities for physical and mental health improvement \cite{1de2013streetscape, 2mcpherson2011million}. It can also contribute to some extent to noise mitigation \cite{3van2016view, 4ferrini2020role}. In addition, urban greenery relates to heat regulation in cities, such as vegetation can reduce the heat island effect through shading and transpiration \cite{11liu2017machine, 6zhang2014cooling}. Therefore, street-level urban greenery is a significant presence for people in terms of aesthetics and a convenient strategy for adaptive environmental design in urban life, creating thermally comfortable and more attractive living environments. Our main goal is to maximize the use of the urban greenery on the street level, and an effective means is to plan the best greenery path, which can improve the quality of life and promote the physical and mental health of urban residents. Moreover, from a novel perspective, to plan the travel routes of travelers \cite{runGVI, 1de2013streetscape}.
    
\begin{figure}[bt!]
\centering
\includegraphics[width=0.8\linewidth]{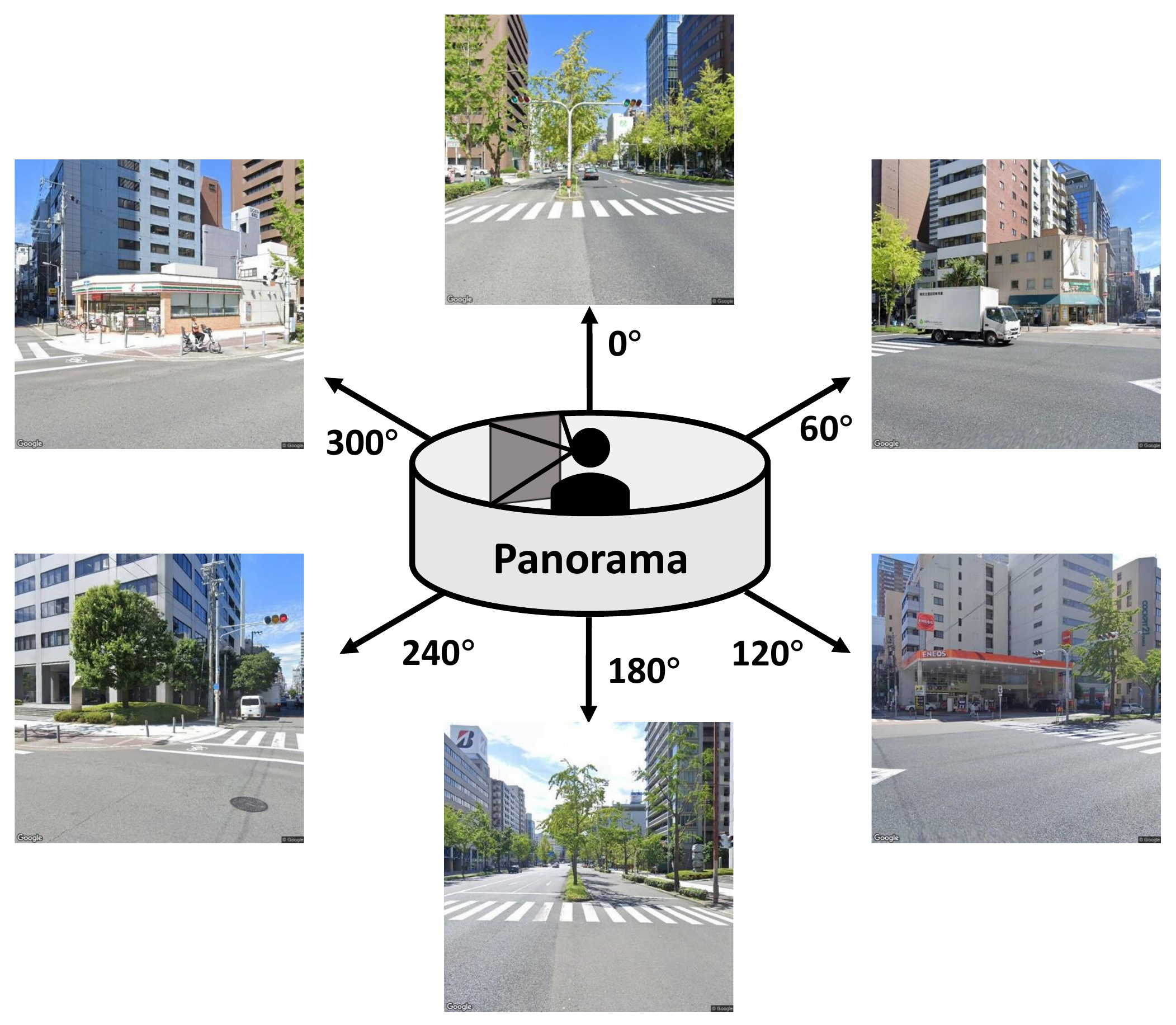}
\caption{The example of obtaining a street view image from a panorama in this paper.}
\label{fig:quanjingtu}
\end{figure}

\begin{figure}
\centering
\subfloat[Street view image]{
\label{fig:svirsi.svia}
\includegraphics[width=0.45\linewidth]{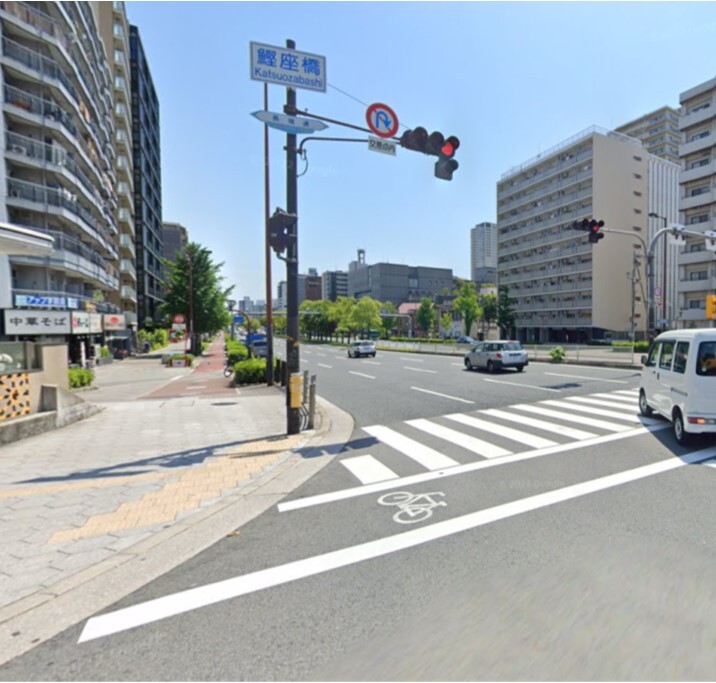}}
\hfill
\subfloat[Aerial optical remote sensing image]{
\label{fig:svirsi.svib}
\includegraphics[width=0.45\linewidth]{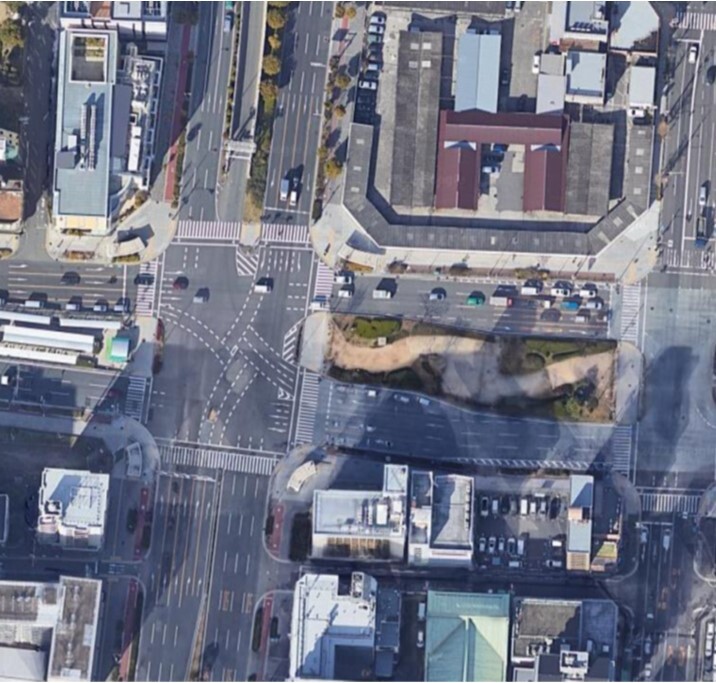}}
\caption{Illustration of \protect\subref{fig:svirsi.svia} street view image and \protect\subref{fig:svirsi.svib} aerial optical remote sensing image of the same location in Google Maps.}
\label{fig:svirsi}
\end{figure}

    Google Street View (GSV) is an efficient method to obtain data resources on urban greenery at the street level. GSV is an interactive web map that is universally accessible and has wide coverage worldwide. It provides a 360$^\circ$ panoramic view of the city, capturing all scenes of a street or neighborhood (see Fig. \ref{fig:quanjingtu}). In addition to street view image (Fig. \ref{fig:svirsi.svia}), remotely sensed data, such as optical remote sensing image (Fig. \ref{fig:svirsi.svib}), are also used as data resources and as a counterpart to street view image in studies of the green index \cite{remoteimage2019, rsistudy2020,streetviewgis2021}. The street view image shows basic information about the landscape from a human perspective, shifting the usual perspective from vertical to horizontal, enabling new insights into the environment and facilitating new applications \cite{streetviewgis2021}, offering numerous possibilities for urban green index studies. Planning the best GVI paths takes what people see on the ground as its starting point, and street view image has unparalleled advantages over most aerial remote sensing methods \cite{li2015assessing}.

    There are many evaluation indexes for the evaluation of street-level urban greenery, among which the current research \cite{li2015assessing, dong2018green} is more widely focused on the Green View Index (GVI) for GSV. The GVI can measure the ratio of greenery within the people's field of view and is more suitable for describing the environment observed by the human eye, which can partially compensate for the shortcomings of traditional assessment indexes \cite{8yang2009can}. With the development of science and technology, there has been tremendous progress from the collection of street view images to image processing \cite{ye2019urban,wang2021noisy}. However, most studies stay at the level of collecting primary GVI data resources \cite{7tong2020evaluating}, focusing on how to use new technology tools to turn the urban landscape into visualization charts and then analyze the distribution characteristics of urban greenery, such as GVI maps, heat maps \cite{li2015assessing}.
    
    Therefore, utilization of the GVI data resources is an effective way to denote street-level urban greenery. Calculating the best GVI path takes advantage of GVI as much as possible. The problem of computing the best path of GVI equates to the longest path problem, which in graph theory and theoretical computer science refers to finding the longest path of length in each graph \cite{longestpth}. In an unweighted graph, the number of edges is the path length, while in a weighted graph, the path length is the sum of the weights. Unlike the shortest path problem, which can be solved in polynomial time, the longest path problem is NP-hard, meaning that unless $P = NP$, corresponding to an arbitrary graph, there is no way to solve the problem in polynomial time \cite{longestpth}. Thus, the longest path problem is a challenging and vital problem. We tried to find the best GVI path by transforming the difficult and complex longest path problem into a shortest path problem by attempting different shortest path algorithms, such as Dijkstra's Algorithm, Floyd-Warshall Algorithm, Bellman-Ford Algorithm \cite{shortestalgo}. Finally, we used the Floyd-Warshall Algorithm to find the best GVI path, which we will explain in detail in section \ref{sec:fwalgo}.

\begin{figure}[bt!]
\centering
\includegraphics[width=1\linewidth]{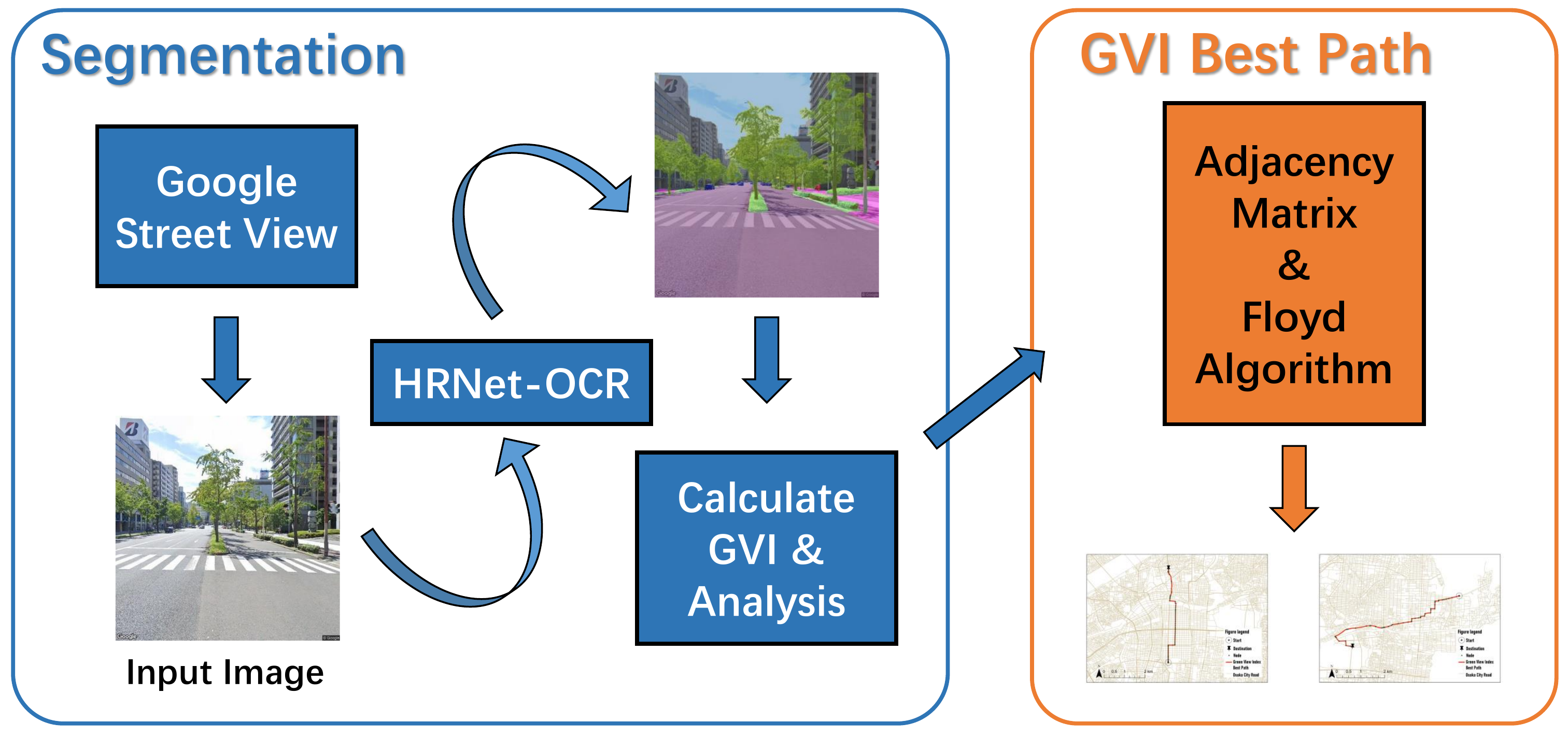}
\caption{The total flowchart of the proposed method.}
\label{fig:flowchart}
\end{figure}

    In this paper, we aim to calculate the GVI data by street view image analysis and use the GVI data to calculate the GVI best paths in Osaka City. The flowchart is shown in Fig. \ref{fig:flowchart}.

    (1) Firstly, the coordinates of all street network road intersections in Osaka City were collected through the open-source Google API (application programming interface) as the investigation site for the study, with a total of 49,770 nodes. We set the width of the view field to 60$^\circ$ to collect more comprehensive street view information. We selected one image at every 60 degrees to get a set of six images containing 0$^\circ$, 60$^\circ$, 120$^\circ$, 180$^\circ$, 240$^\circ$, and 300$^\circ$, which can cover 360-degree panoramic street view. A total of 298,620 street view images of Osaka City were collected. 
    
    (2) The obtained images are segmented by the image segmentation model HRNet-OCR \cite{HRNet-OCR}. The value of GVI, the ratio of the green view index (vegetation and terrain) in the landscape elements of the street view image, was calculated to assess the degree of green view of the image. 
    
    (3) Create a map of the GVI distribution of Osaka City. The evaluation criterion for the GVI is referred to in the survey report of the Ministry of Land, Infrastructure, Transport, and Tourism of Japan\footnote{\url{https://www.mlit.go.jp/kisha/kisha05/04/040812_3/01.pdf}} and the satisfaction criteria of the first phase of the Kyoto Greening Promotion Plan\footnote{\url{https://www.city.kyoto.lg.jp/kensetu/cmsfiles/contents/0000102/102008/planhonpen.pdf}} about the GVI. The GVI was divided into four grades: 0$\sim$10\%, 10$\sim$18\%, 18$\sim$25\%, and 25\% or more. Based on this criterion, a general overview of GVI distribution and degree of satisfaction in Osaka City was constructed. 
    
    (4) We have adopted a general approach to convert the GVI best path problem into a shortest path problem. Although the algorithm for the shortest path exists in ArcGIS, it does not correctly calculate the best GVI path in complex scenarios.

    The following contributions were achieved:
\begin{enumerate}
\item A generic approach is proposed to achieve the best path of GVI in complex situations, breaking through the limitations of ArcGIS software and visualizing it on a map with geographic data.
\item We propose a realistic scenario that uses the GVI in the forward direction of the route instead of using each node's average value and calculating the best GVI path in an ideal area (Imazu Park).
\end{enumerate}

\section{Related Works}
    \subsection{Google Street View and Green View Index}
    Street view image has been used extensively in various types of research \cite{zjxmdpi,lyq2022, zjxieee}. These studies have shown that street view image datasets are useful and quantitative tools to help policymakers, planners, and researchers to understand the landscape from the human perspective. In recent years, Google Maps\footnote {\url{https://www.google.com/streetview/}} and Baidu Maps\footnote {\url{http://lbsyun.baidu.com/}} have provided Point of Interest (POI) alternatives in their public Geographic Information System (GIS) databases. With the development of these public GIS databases, easier access to raw data (street view image API) facilitates large-scale street view image-based research \cite{11liu2017machine,16kang2018building}.

    Many previous studies have been conducted on urban street greenery using Google Street View. \cite{2.1.1} demonstrated that urban green spaces could be used for recreational sports and thus promote the health of residents, showing that street greenery has a positive impact on health. \cite{2.1.2} demonstrated that the physical and mental health benefits of physical activity in green spaces might exceed those of physical activity in other environments. \cite{gviphsical} verified that street greenery, in addition to green landscapes in parks, has a positive effect on residents' frequency of physical activity. The findings also reveal the impact of eye-level street greenery on residents' physical activity levels, thus contributing to developing and implementing healthy cities to stimulate physical activity. In \cite{walk}, from the starting point of analyzing the correlation between walking time and green landscape, it demonstrated that GVI has a stronger correlation with walking time compared to traditional greenery variables. The results of this study provide valuable guidelines for policy directions to construct and promote pedestrian walking environments.

    \subsection{Semantic Segmentation for GVI Analysis}
    In the computer vision field, deep learning uses Deep Neural Networks (DNN) for feature extraction and parameter optimization \cite{DNN, ma2022building}. Semantic segmentation is one of the branches to classify each pixel in an image with different classes based on training data. It is a kind of supervised learning \cite{23zhao2017pyramid}. For natural images, it is achieved by using neural networks to identify landscape categories at the pixel level rather than recognizing their type and position in the overall image. There are many scenarios where semantic segmentation is currently used, for example, in civil engineering \cite{17zhang2019concrete}, to detect cracks in concrete; in the medical field to identify diseases in X-ray images \cite{wang2021automatic} and in autonomous driving for road boundary and object detection \cite{19siam2017deep}.

    Models for semantic segmentation have developed rapidly in recent years, and representative models include U-Net \cite{20ronneberger2015u}, SegNet \cite{21badrinarayanan2017segnet}, DeepLabv3+ \cite{deeplabv3+}, PSPNet \cite{23zhao2017pyramid}, and so on. All these models have shown satisfactory results regarding a common evaluation metric, intersection-over-union (IoU). In this study, We strongly refer to the official model benchmark rankings of pixel-level semantic labeling task of the Cityscapes Dataset \cite{cordts2016cityscapes}, and we decided to use HRNet \cite{HRNet} as the backbone and OCRNet as the semantic segmentation model, to analyze street view images and calculate GVI.
    
    Several studies have proposed using semantic segmentation to analyze urban green views and landscapes. \cite{newyork} used Google Street View images at different times to map and analyze the spatial distribution and temporal variation of the GVI in New York City. \cite{yokohama} proposed an improved GVI method to analyze the central neighborhoods of Yokohama, Japan. Besides, many pioneering studies have existed on the integration of landscape data processing with GIS research, such as \cite{25cetin2015using} which used geographic data analysis tools of ArcGIS to analyze and assess the accessibility of urban green spaces. In \cite{26yamagata2016value}, aerial photographs were implemented on GIS for view area analysis. Three types of landscapes were quantified: open landscapes (visibility), green landscapes (visibility of open spaces), and marine landscapes (visibility of the ocean). It indicated that ArcGIS software is a handy tool for integrating photo information quantification and geographic data \cite{26yamagata2016value}. To the best of our knowledge, no research has been published to analyze the GVI of Osaka City using an open-source street view API, so we decided to fill this gap and perform a systematic GVI analysis of Osaka City.
    
    \subsection{The Studies of Path Planning}
    Many studies tend to develop new routing algorithms for finding routes with optimized exposure. These studies tend to examine multiple types of exposure factors and plan for optimal routes \cite{greenpath, omsgreen, 2.3.1}. These exposure factors include green landscape, air quality, and noise index, which also include studies that analyze individual exposure factors \cite{heatstress}. \cite{runGVI} analyzed the effect of GVI on runner satisfaction. A positive correlation was found between running satisfaction and nature exposure as assessed by the Normalized Difference Vegetation Index (NDVI) and the GVI. \cite{greenpath} developed a software Green Paths using python toolkits to analyze paths in Helsinki under different exposure scenarios, such as traffic noise levels, air quality, and street greenery. It employs a novel impedance function on the path to achieve path planning with different exposures. \cite{omsgreen} proposed a system for generating customized pedestrian routes based entirely on OpenStreetMap (OSM) data. By selecting OSM features, tags considered green features are extracted, and the green areas are represented in the top view. \cite{2.3.1} designed healthy routes considering factors such as noise and air pollution, which can have a positive effect on the sustainability of future cities. Using a healthy route planner can minimize the impact of pollution on all citizens, especially on the most vulnerable groups of the population, and reduce potential health problems and associated costs. \cite{hernandez2018allergyless} proposed a recommendation system that addresses citizens' travel problems by informing them which walking routes minimize allergen exposure. Allergen data are collected through contamination monitoring points, allergen predictions are made for areas without monitoring points, and routes to avoid allergens are planned and validated. \cite{heatstress} presented a two-step approach for planning routes to reduce individual heat stress for cities or areas where walking or bicycling are valid options, demonstrating that heat stress can be reduced in most cases.

\begin{figure}[bt!]
\centering
\includegraphics[width=1\linewidth]{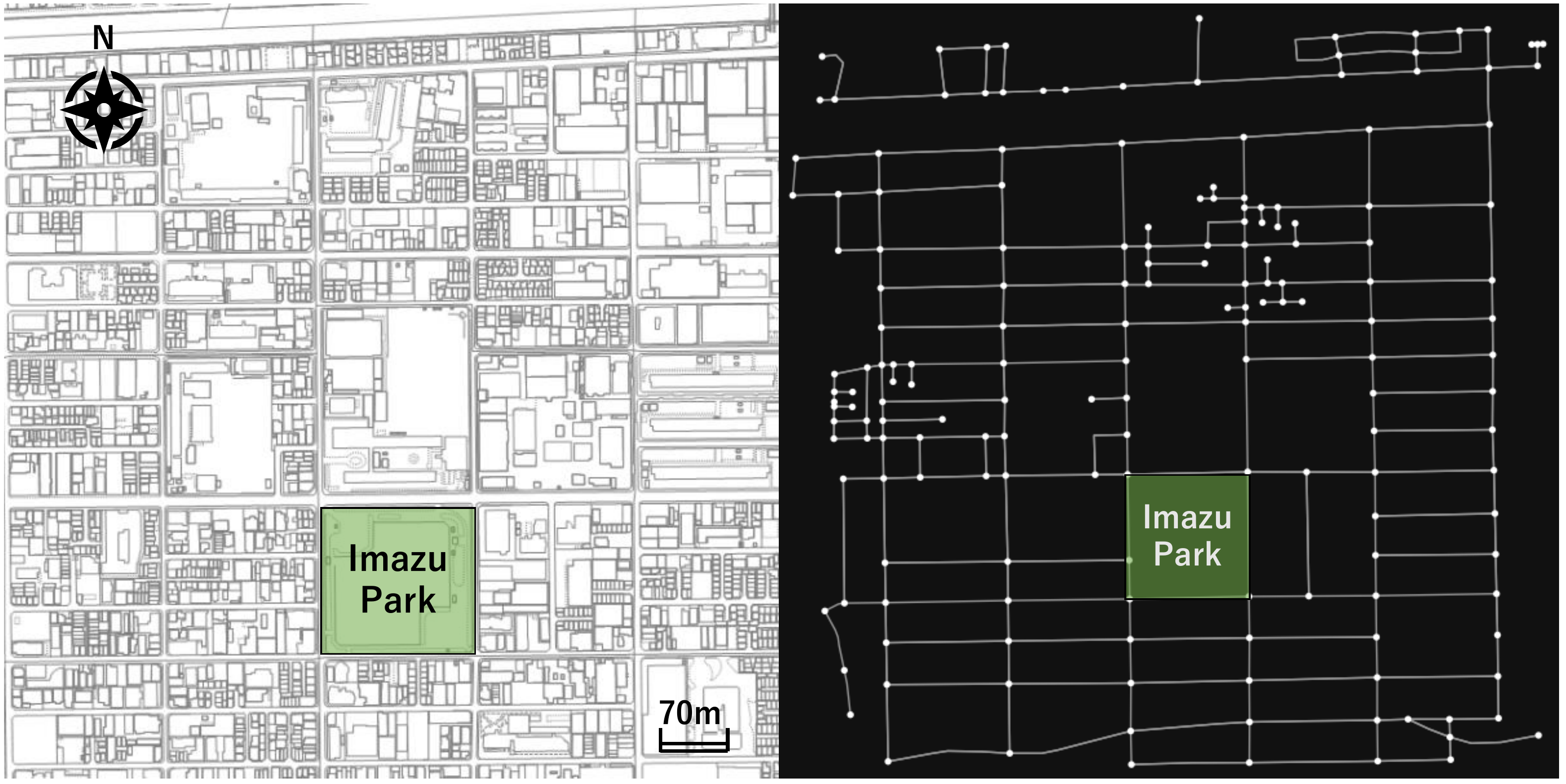}
\caption{The diagram of Imazu Park and surrounding area.}
\label{fig:imazu}
\end{figure}

\section{Material and Method in Deep Learning}
    In this study, we constructed a GVI distribution map system according to the following process:
    (1) Obtain the street view images of Osaka City.
    (2) Semantic segmentation of street view images with HRNet-OCR, with HRNet as the backbone and OCRNet as the semantic segmentation model. 
    (3) Calculate each node's GVI in different directions and the average GVI.
    (4) Integrate the geographic data with GVI using ArcGIS to visualize the GVI.
    (5) The proposed method calculated and visualized the GVI best path with geographic data.

\subsection{Research Area}
    We considered the entire city of Osaka as the study area since we used it to verify that our proposed algorithm can calculate the best GVI path for any two nodes within the city of Osaka.

    Osaka City in Japan is the administrative, economic, cultural, and transportation center of the Kinki region and western Japan. It has an area of approximately 225.21 km$^2$ and comprises 24 administrative districts. The Osaka Metropolitan Area and the Keihan-Kobe Metropolitan Area are formed, with Osaka City as the center. The Keihan-Kobe Metropolitan Area is second only to the Tokyo Metropolitan Area in terms of Gross Domestic Product (GDP) in Japan and ranks among the highest in the world. It was ranked 35$^{th}$ in the world in the "Global Cities Index 2020" ranking of world cities by a U.S. think tank\footnote{\url{https://www.kearney.com/global-cities/2020}}. It is well suited as a target city for researching urban landscapes. Therefore, we set Osaka City as the first target area to calculate and visualize GVI distribution to understand the city overall. The visualization of the distribution facilitates a more intuitive understanding.
    
    In the section on GVI best path, it is unpractical to analyze the GVI path of all of Osaka City because we could not get a street view facing the road at the intersection in Osaka City by Google Map APIs. Therefore, we select a representative area in Osaka City to verify whether the best GVI path methods are feasible. In this area, it is possible to get an ideal angle of view facing due east, west, north, and south at each intersection, i.e., the direction of view is parallel to the road. Therefore we set Imazu Park and its surrounding area (see Fig. \ref{fig:imazu}) as the second target area to analyze the detailed GVI best path.

\subsection{Get Street View Images of Osaka City}
    We used the Python OSMnx package \cite{28boeing2017osmnx} to obtain the coordinate data of all intersections in Osaka City from the Open Street Map to get map networks of Osaka City. With the overall networks, we can get coordinates of all the intersections. In recent years, companies such as Google, Amazon, and Twitter have been actively providing data through web service APIs to leverage their various types of big data. Google Street View Image API makes downloading Google Street View images easy. Therefore, we collected street view images by Google Maps API with obtained coordinates of all intersections in Osaka City (a total of 49,770 nodes). Street view images were collected from six angles (0$^\circ$, 60$^\circ$, 120$^\circ$, 180$^\circ$, 240$^\circ$, 300$^\circ$) in the direction of the street centered at each intersection. Their size was 640$\times$640 because, by Google Maps API, we can only get the maximum size of 640$\times$640. We need as much as possible to get a high-resolution input image and a more accurate result by HRNet-OCR. We download 298,620 images of Osaka City. One thing that needs to be mentioned is that the street view images obtained may not be suitable for analysis, such as night-time street scenes, indoor images, and blurred images. These nodes need to be manually filtered and discarded.

\subsection{Semantic Segmentation and Calculation of GVI}
    Semantic segmentation, a powerful technique in computer vision, is efficient and accurate since it can classify pixels in an image into different classes with high accuracy on large datasets \cite{streetviewgis2021}. Semantic segmentation not only quantifies the distribution of GVI but also applies to large-scale datasets, making it possible to analyze the urban landscape on a large scale. In this study, we use HRNet-OCR as the backbone model for GVI extraction, which was pre-trained on Cityscapes to generate segmented images into 19 landscape elements. The motivation for choosing such a method for semantic segmentation is that the HRNet has meaningful semantic features and the OCR method explicitly transforms the pixel classification problem into an object region classification problem \cite{OCRNet}. 

\begin{figure}[bt!]
\centering
\includegraphics[width=1\linewidth]{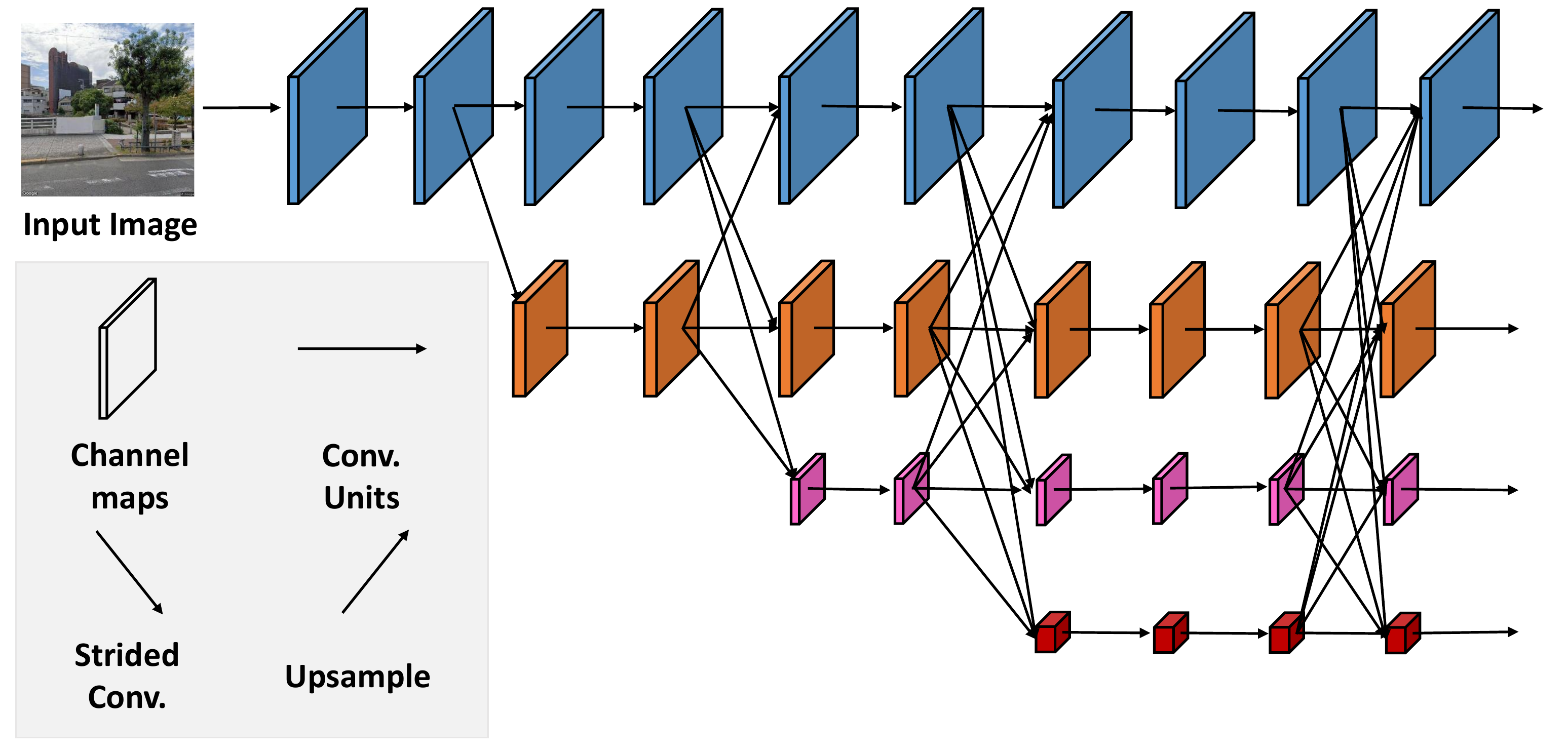}
\caption{The basic structure of HRNet backbone.}
\label{fig:HRNet}
\end{figure}

    Semantic segmentation is location-sensitive in tasks of computer vision domain. Maintaining a high-resolution feature map is a simple and effective way to process the model to make the task location information more accurate.
    HRNet achieves both complete semantic information and accurate location information by parallelizing multiple branches of the resolution, coupled with the constant interaction of information between different branches \cite{HRNet}. The basic structure of HRNet backbone is shown in Fig. \ref{fig:HRNet}. The main idea of OCR is consistent with the original definition of the semantic segmentation problem, i.e., the class of each pixel is the class of the object to which the pixel belongs. In other words, the main difference between the contextual information of PSPNet and DeepLabv3+ is that the OCR method explicitly enhances the object information. The pipeline of OCRNet is shown in Fig. \ref{fig:OCRNet}.

\begin{figure}[bt!]
\centering
\includegraphics[width=1\linewidth]{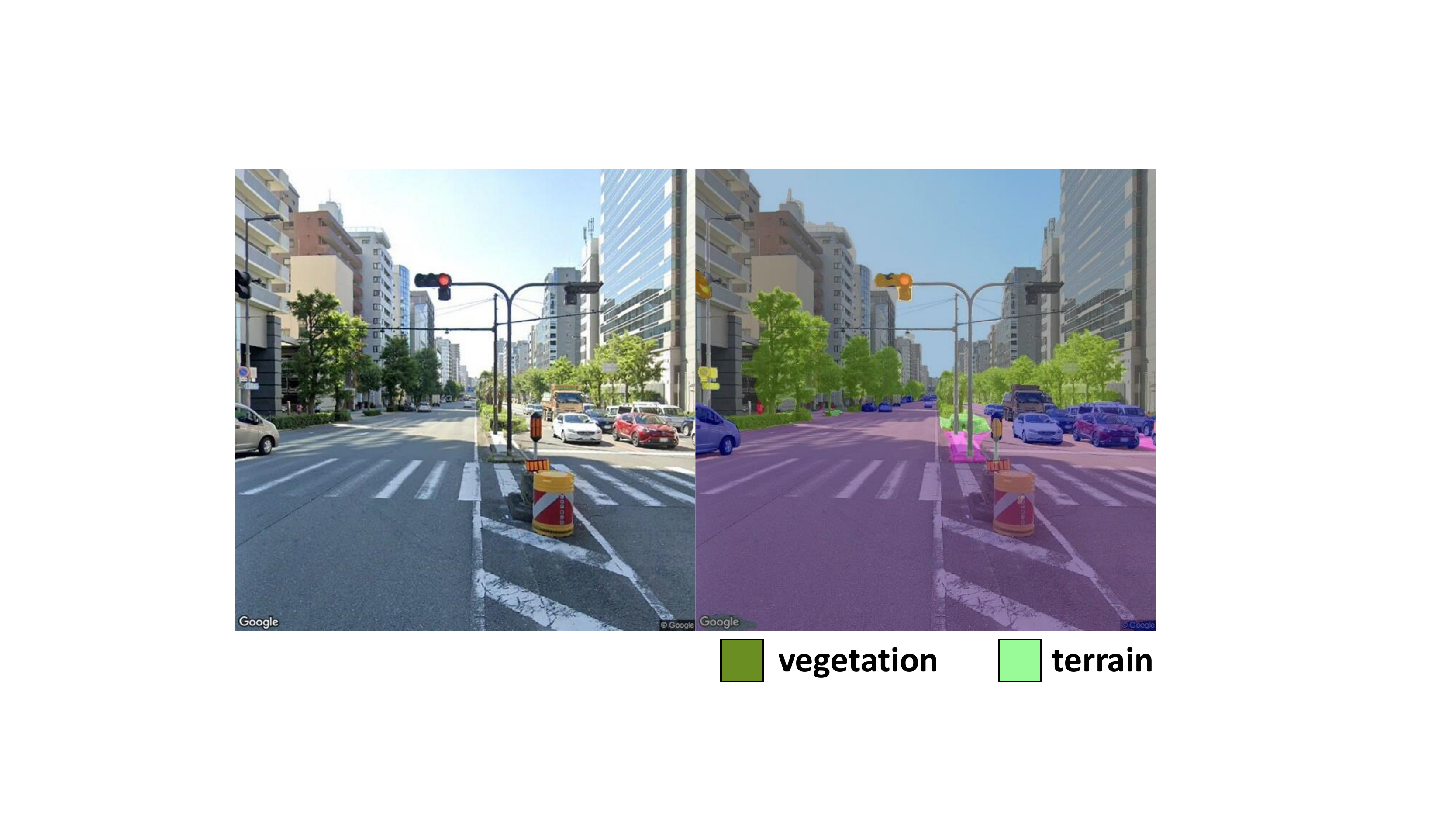}
\caption{Inference example of street view image of Osaka City and its corresponding labels (vegetation and terrain) we used.}
\label{fig:ss}
\end{figure}

    For the training process, we downloaded the labeled open-source dataset Cityscapes, which contains 5,000 images with the correlated fine label in Europe. The entire dataset is divided into three parts, 75\% of them are for training, 10\% for validation, and 15\% for the test. Even though the elemental composition and details are different, for the composition at the class level, we can get precise segmentation results by HRNet-OCR, which is pre-trained on Cityscapes. The colored label image is transformed into a true label image which uses the class index as a pixel number to improve training efficiency. Because the original training data has a large image size, we also need to set the crop window with a relatively small size on the local part of the original image, ensuring not to miss much basic information if we straight resize the original image in the training dataset. We use Intersection-over-Union (IoU) to evaluate the segmentation result for the evaluation criterion of training results. It consists of three parts. For each class, IoU calculates the generated region of one class with the label region. The IoU is defined as below:
    
\begin{figure}[bt!]
\centering
\includegraphics[width=1\linewidth]{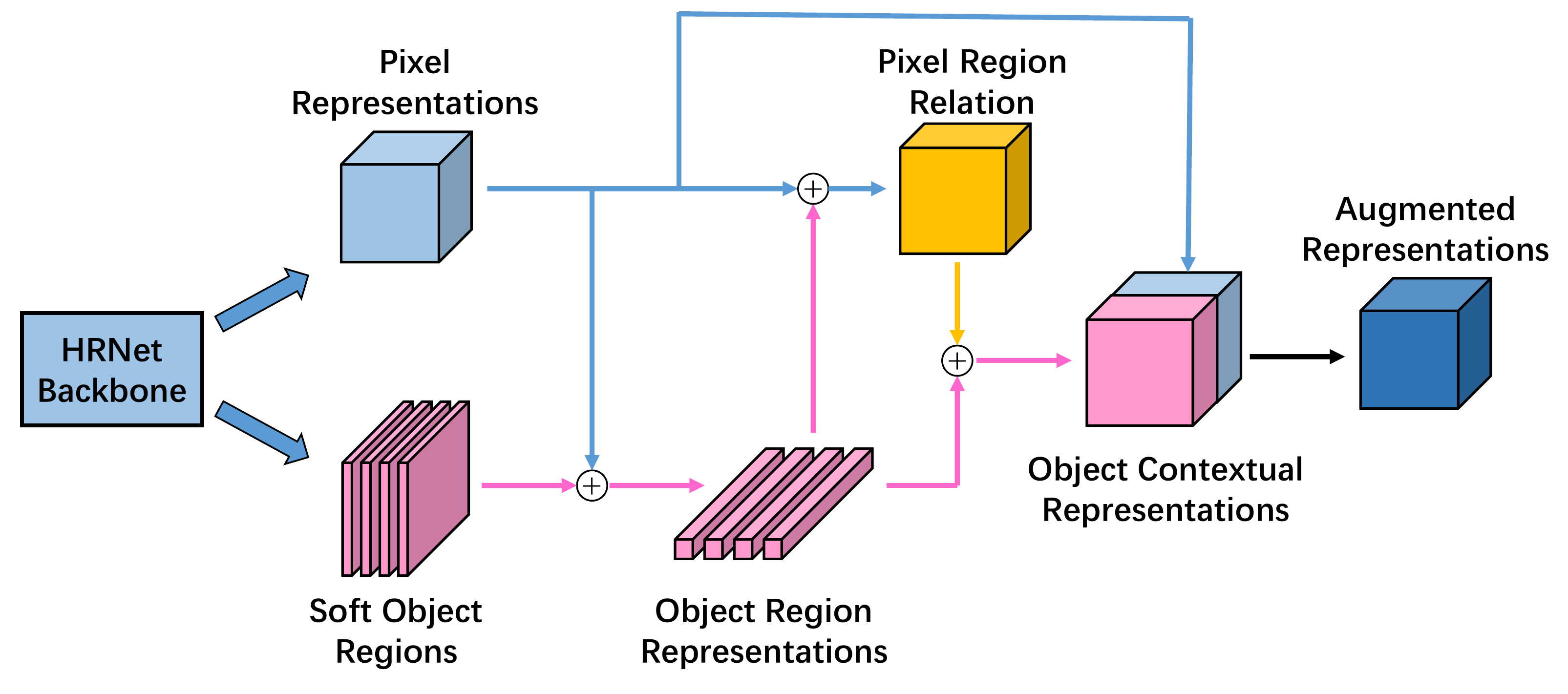}
\caption{The pipeline of OCRNet \cite{OCRNet}.}
\label{fig:OCRNet}
\end{figure}

\begin{equation}
    IoU=\frac{TP}{TP+FP+FN}
\end{equation}

where ${TP}$ is True Positive, $FP$ is False Positive, and $FN$ means False Negative. This criterion can be simplified by the ratio of the intersection area of each class between the generated region and label region to the union area between them. HRNet-OCR achieved the mean IoU to 80.6\% in our test set. The inference result of Osaka City is shown in Fig. \ref{fig:ss}. The size of the generated segmented image is closely related to the generation efficiency. As Osaka City contains numerous images, we have reduced the generated image size from 2048$\times$1024 to the original image size, which is 640$\times$640, to improve the inference efficiency of segmentation. We spent almost 90 hours generating segmented street view images of Osaka City and selecting vegetation and terrain as target colors to calculate GVI.

In a segmented image, for each angle $i\in$ (0$^\circ$, 60$^\circ$, 120$^\circ$, 180$^\circ$, 240$^\circ$), $G_i$ is the number of greenery pixels. $T_i$ means the total number of this segmented image. $GVI_i\in\left\{{0,1}\right\}$ means the corresponding GVI of this angle, which is given by

\begin{equation}
    GVI_i=\frac{G_i}{T_i}\times100\%
\end{equation}

Each intersection node of $M$ contains the number of $m$ street view images. The average of $GVI_i$ represents the GVI of this node. Therefore, the $GVI_{avg}$ of each intersection is given by

\begin{equation}
    GVI_{avg}=\frac{1}{m}\sum_{i=1}^{m}GVI_i\times100\%
\end{equation}

where $m=6$ in this paper.

\begin{figure*}
\centering
\subfloat[The GVI distribution of node map in Osaka City.]{
\label{map.sub.1}
\includegraphics[width=0.45\linewidth]{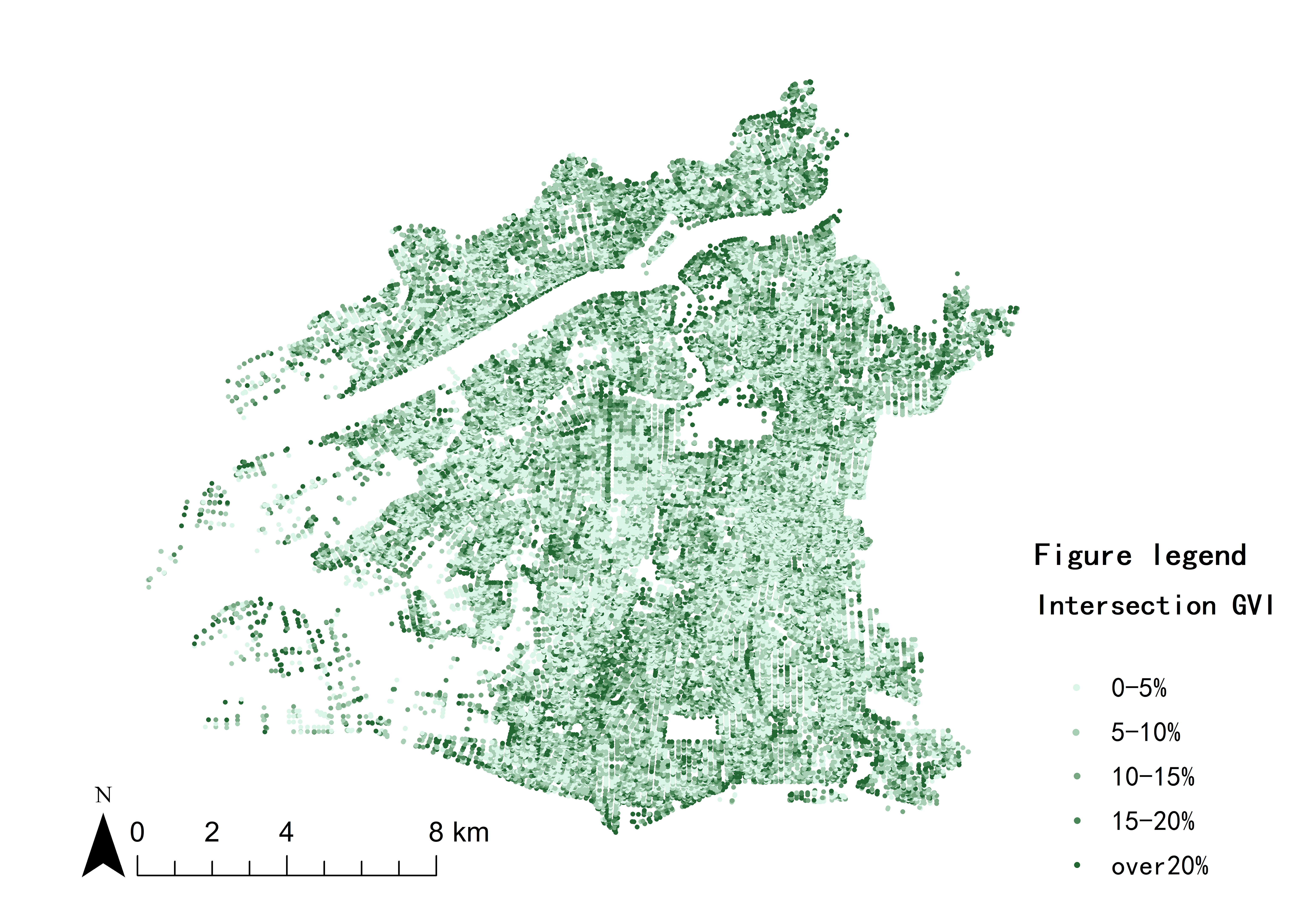}}
\hfill
\subfloat[The GVI distribution of line map in Osaka City.]{
\label{map.sub.2}
\includegraphics[width=0.45\linewidth]{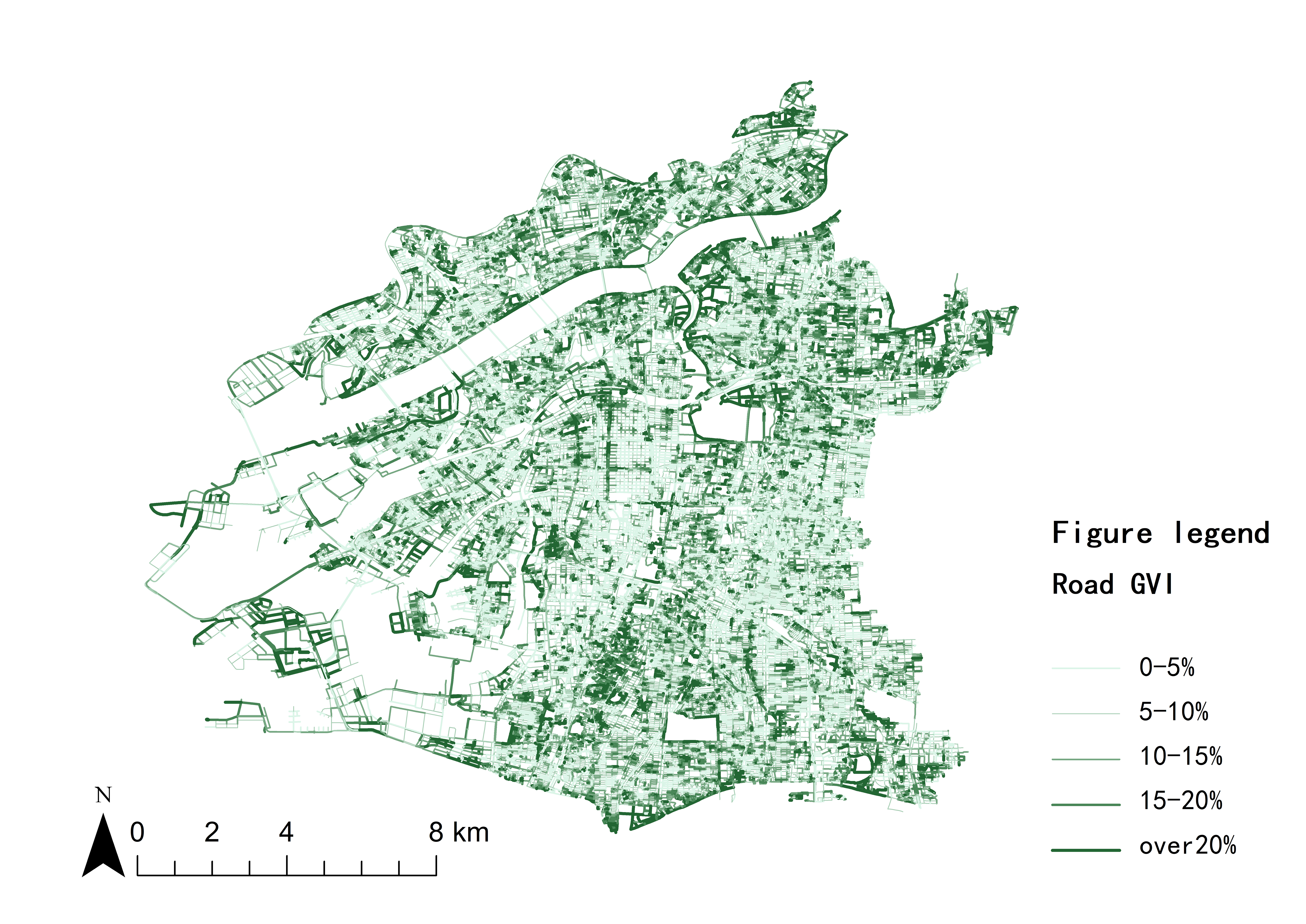}}
\caption{The node map and line map of GVI distribution in Osaka City.}
\label{map}
\end{figure*}

\subsection{Integration and Visualization of GVI data}
After getting the geographic data, including street line and node network by open street map, we can also get an excel with GVI data and coordinate. We use ArcGIS to integrate them with node and line maps to manually visualize the GVI distribution in Osaka City. The created node map of the distribution of GVI in Osaka City is shown in Fig. \ref{map.sub.1}. Each node is represented by GVI${_{avg}}$ and coordinates with various color depths in five classes.

Based on the GVI node map, we adopted an assignment method, which uses the average GVI of the vertices on both sides of the street to represent the value of the overall street. Therefore, the line map of GVI distribution in Osaka City was obtained as shown in Fig. \ref{map.sub.2}.

\begin{figure}[bt!]
\centering
\subfloat[]{
\label{error.sub.1}
\includegraphics[width=0.3\linewidth]{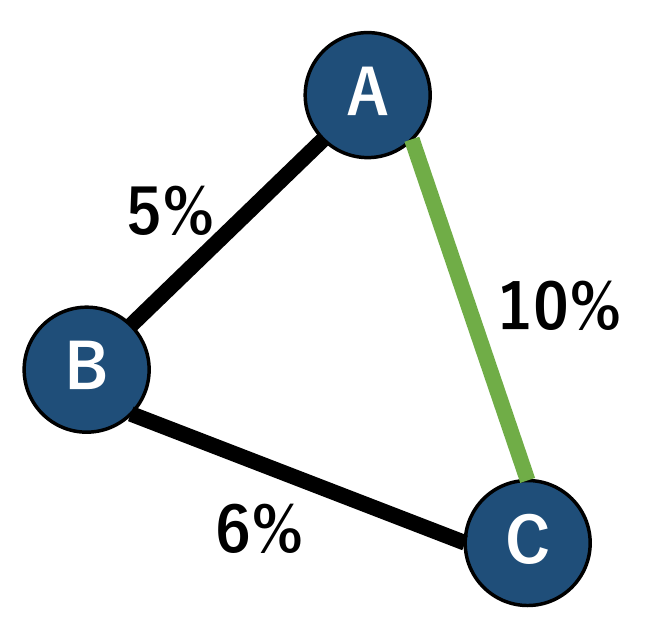}}
\quad
\hspace{6mm}
\subfloat[]{
\label{error.sub.2}
\includegraphics[width=0.3\linewidth]{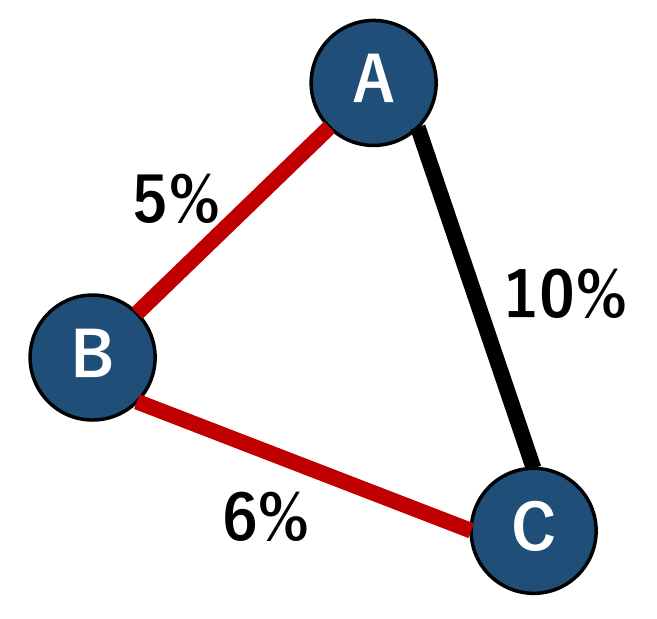}}
\caption{Define the starting point as A; the destination is C. The value (\%) means the GVI of each line. \protect\subref{error.sub.1} The error path (green) by ArcGIS software. \protect\subref{error.sub.2} The correct path (red).}
\label{error}
\end{figure}

\section{GVI Best Path Analysis}
\subsection{GVI Best Path by ArcGIS -- An Unfeasible Approach}
    \subsubsection{Attempt by cell statistics tool}
    Firstly, by the cell statistics tool, we can get the raster returned by the corridor analysis tool, and the sum of the cost distances (cumulative cost) of the two input cost rasters is calculated for each location of the image. With this idea, the line graphs of the GVI distribution of Osaka City were exported to raster files according to the satisfaction thresholds of GVI (0$\sim$10\%, 10$\sim$18\%, 18$\sim$25\%, and more than 25\%) and assigned resistance values of 200, 150, 100, 50 respectively. The land use data of Osaka City was also assigned with the corresponding resistance values according to the land attributes. We tried to assign a resistance value greater than 200 to all land outside the road and make the path travel in the direction of the road. We assigned Nagai Park as the starting point and Utsubo Park as the destination as an example. The Nagai Park was used as a starting point; therefore, two cost accumulation rasters were created. We took the same approach to set Utsubo Park as another starting point, the process of creating cost surfaces based on individual locations occurs at each location of the input raster, and the total cumulative cost of the path through the image is calculated. However, the result we obtained was a very coarse path band, which had no practical value for the GVI best path analysis, so we decisively abandoned this method.

\begin{algorithm} 
    \renewcommand{\algorithmicrequire}{\textbf{Input:}}
	\renewcommand{\algorithmicensure}{\textbf{Output:}}
	\caption{Floyd Algorithm using Adjacency Matrix.} 
	\label{algo1} 
	\begin{algorithmic}
		\REQUIRE Adjacency Matrix $graph$, number of nodes $N$, the matrix used to store the parent nodes $parents$. 
		\FOR {$k$ in $N$times}
		\FOR {$i$ in $N$times}
		\FOR {$j$ in $N$times}
		\IF{$graph[i][k] + graph[k][j] < graph[i][j]$}
		\STATE $graph[i][j] = graph[i][k] + graph[k][j]$
        \STATE $parents[i][j] = parents[k][j]$
        \ENDIF
        \ENDFOR
        \ENDFOR
        \ENDFOR
		\ENSURE a $parents$ matrix with the shortest path.
	\end{algorithmic} 
\end{algorithm}    
    
\begin{algorithm} 
    \renewcommand{\algorithmicrequire}{\textbf{Input:}}
	\renewcommand{\algorithmicensure}{\textbf{Output:}}
	\caption{Get adjacency matrix from adjacency table} 
	\label{algo2} 
	\begin{algorithmic}
    	\REQUIRE adjacency table (node, node, GVI) file $data$, number of nodes $N$, (N, N)matrix with diagonal elements as 0 and other elements as inf $graph$
		\FOR {$u, v, c$ in $data$}
		\STATE $graph[u][v] \gets 0$
		\STATE $graph[v][u] \gets 0$ \COMMENT{undirected graph}
        \ENDFOR
		\ENSURE adjacency matrix $graph$ 
	\end{algorithmic} 
\end{algorithm}

\begin{figure*}
\centering
\includegraphics[width=0.9\linewidth]{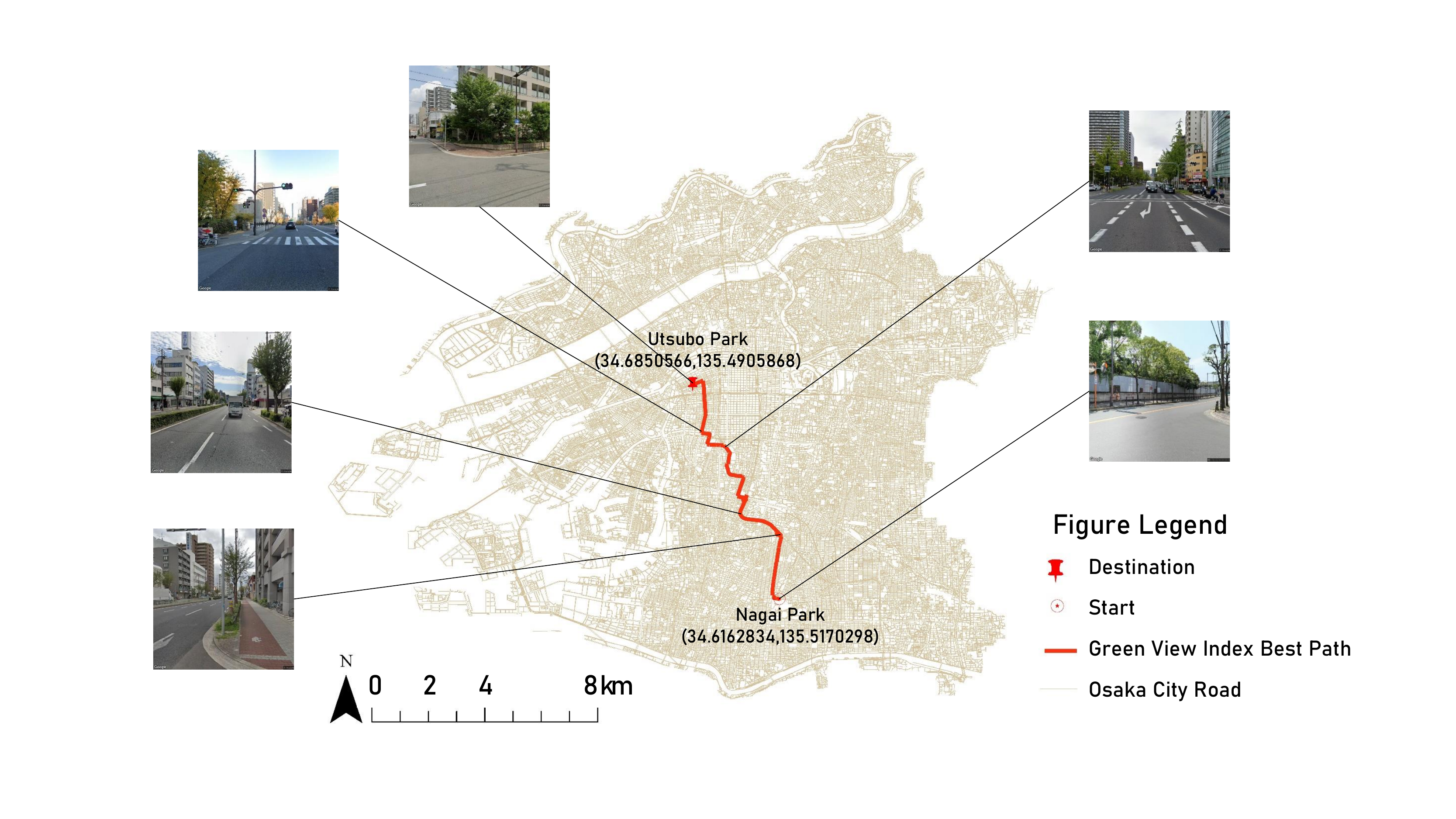}
\caption{The best GVI path from Nagai Park (34.6162834, 135.5170298) \protect\footnotemark  to Utsubo Park (34.6850566, 135.4905868) and street view examples by adjacency matrix and Floyd-Warshall Algorithm.}
\label{fig:Floyd}
\end{figure*}

    \subsubsection{Attempt by geometric network analysis}
    We tried a new method to attempt the calculation of the GVI best path in ArcGIS software. Geometric network analysis has many applications in calculating the shortest distance between two nodes, such as Dijkstra's Algorithm. We tried to see whether this logic could also be applied to try to calculate the GVI best between two nodes. First, we import the GVI node map and GVI line map of the GVI distribution in Osaka City and transform them into a dataset to generate a geometric network. We have the average GVI of each node and set the GVI of the line between two nodes to the average GVI of the two nodes. Because there is no algorithm for the longest path in ArcGIS, we need to adjust the GVI of each line segment to inverse to get the new GVI of this line segment. However, if we take negative values directly, the shortest path problem does not apply (the shortest path algorithm of ArcGIS requires positive weights between nodes). We attempted to convert the maximum GVI problem into a shortest path problem by assigning a value of ($100-GVI$) to each line. In this case, we need the number of nodes to ensure that the result obtained using the shortest path algorithm is correct after the ($100-GVI$) transformation. In ArcGIS software, the lack of the number of nodes results in an incorrect best path towards a path with fewer nodes. This scenario can be seen in the simple dotted line diagram Fig. \ref{error}.

    \subsubsection{Disadvantages}
    The best GVI path does not have the exact attributes as the geometric space, such as the path length, as the shortest path problem, which causes an assortment of problems. For instance, in a long stretch of road in geometric space, where the GVI is exceptionally low, we must incorporate information about the number of nodes to make the best GVI path obtained after the ($100-GVI$) transformation of the graph the correct one. The methods of using ArcGIS software to solve for the best GVI path have huge drawbacks below. 
\begin{enumerate}
\item By the cell statistics tool, a coarse path band is obtained rather than a practical and fine path, which does not apply to the practical situation.
\item ArcGIS software cannot derive the number of nodes passed from the shortest path problem, which leads to an incorrect result.
\item Besides calculating the best GVI path, the simple shortest path problem could not be addressed without the number of nodes passed.
\end{enumerate}
\footnotetext{The two numbers indicate (latitude, longitude).}

\subsection{Our Approach\label{sec:fwalgo} -- A Generic Approach}
    \subsubsection{The best GVI path by adjacency matrix and Floyd-Warshall Algorithm}
    We decided to use the adjacency matrix as the first step and the Floyd-Warshall Algorithm (see Algorithm \ref{algo1}) to calculate the best GVI path between any two nodes. The adjacency relationship between nodes and lines can be obtained from ArcGIS. The adjacency table of all nodes and lines can be acquired by transformation, and the weights are the transformed GVI ($100-GVI$), and the undirected graph is used to represent the adjacency matrix. Since the adjacency matrix contains information about the number of nodes, we can use ($100-GVI$) to convert the problem into a shortest path problem without any concern. As shown in Algorithm \ref{algo2}, with the adjacency matrix, Floyd-Warshall Algorithm can derive a list of node arrangements for the best path, and finally, we can represent the best GVI path on the graph. It is important to mention that we can obtain the adjacency matrix of the region after a long time. For instance, if we use a region containing about 5000 nodes for the adjacency matrix generation operation, it takes nearly three hours to achieve this process. Once we have the adjacency matrix, specifying the start and destination can promptly lead to a GVI best path and import the basic information into ArcGIS to visualize the path.

\begin{figure*}
\vspace{-5mm}
\centering
\subfloat[The GVI best path from Namba (34.667234, 135.500485) to Umeda (34.706838, 135.500434). The average GVI of this path is 10.05\%.]{
\label{map1.sub.1}
\includegraphics[width=0.46\linewidth]{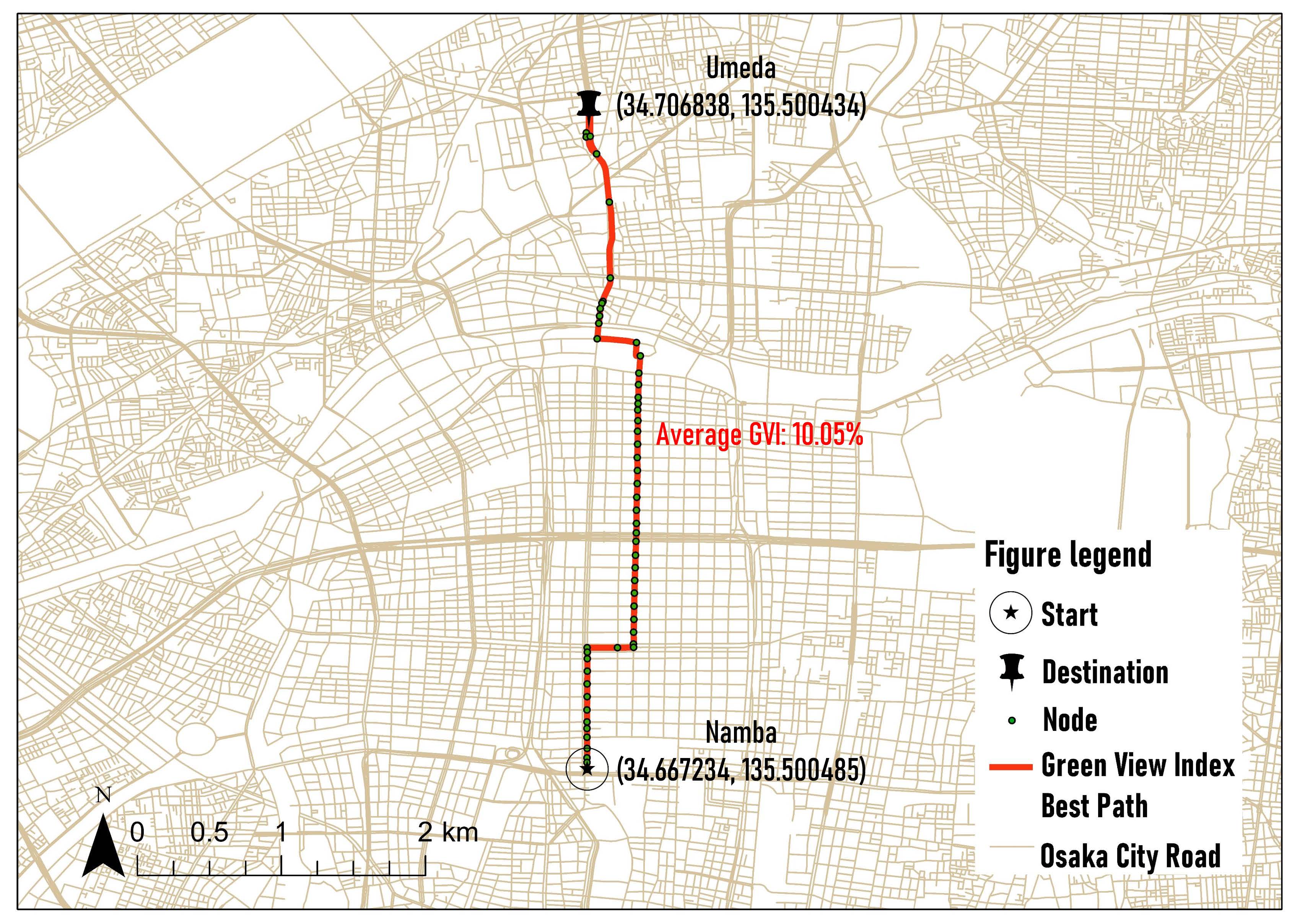}}
\vspace{2mm}
\hspace{4mm}
\subfloat[The GVI best path from Umeda (34.706838, 135.500434) to Namba (34.667234, 135.500485). This path is the same GVI path as \protect\subref{map1.sub.1}.]{
\label{map1.sub.2}
\includegraphics[width=0.46\linewidth]{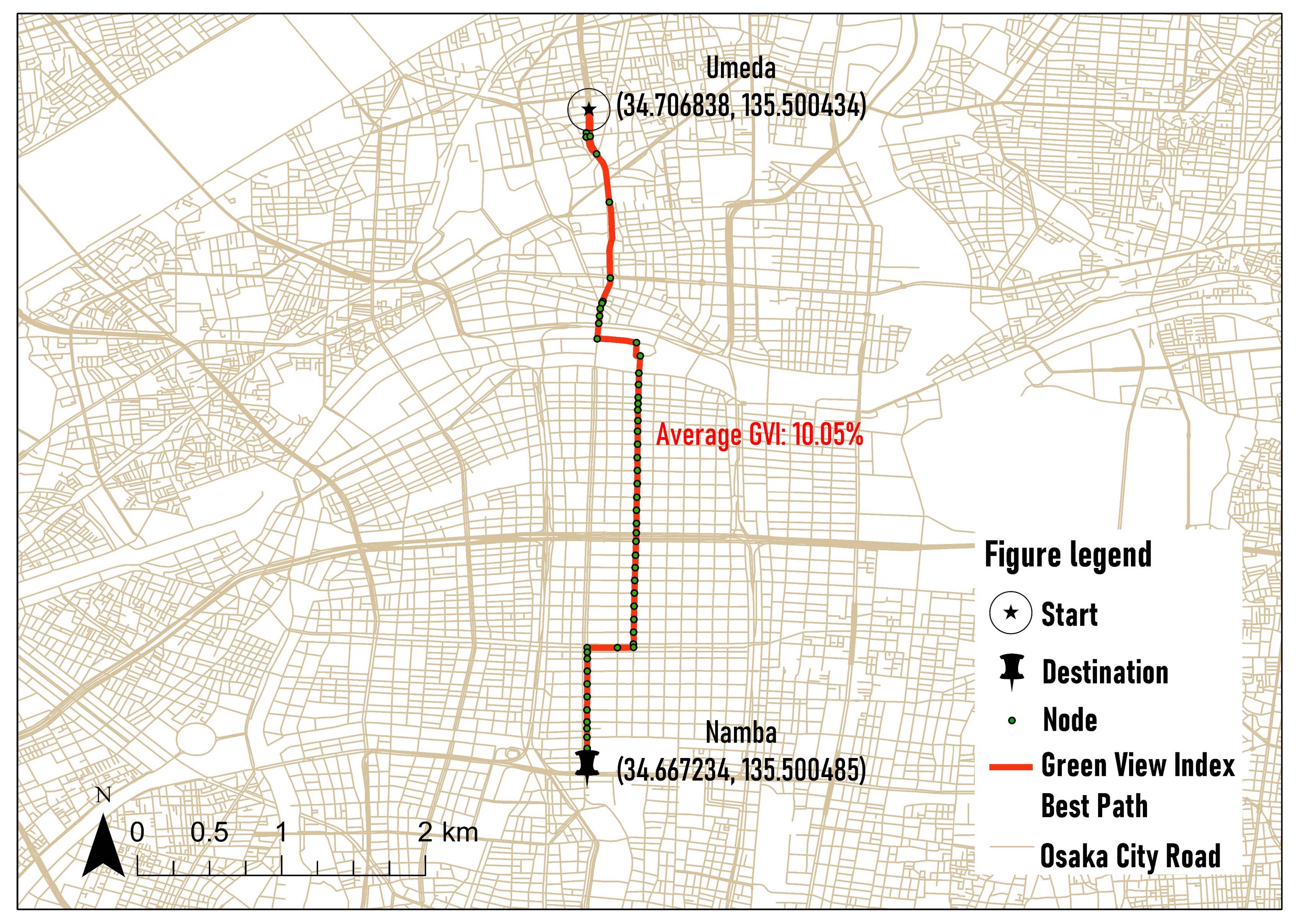}}
\vspace{2mm}
\subfloat[The GVI best path from Suminoe Park (34.609184, 135.473031) to Nagai Park (34.616133, 135.518644). The average GVI of this path is 9.06\%.]{
\label{map1.sub.3}
\includegraphics[width=0.46\linewidth]{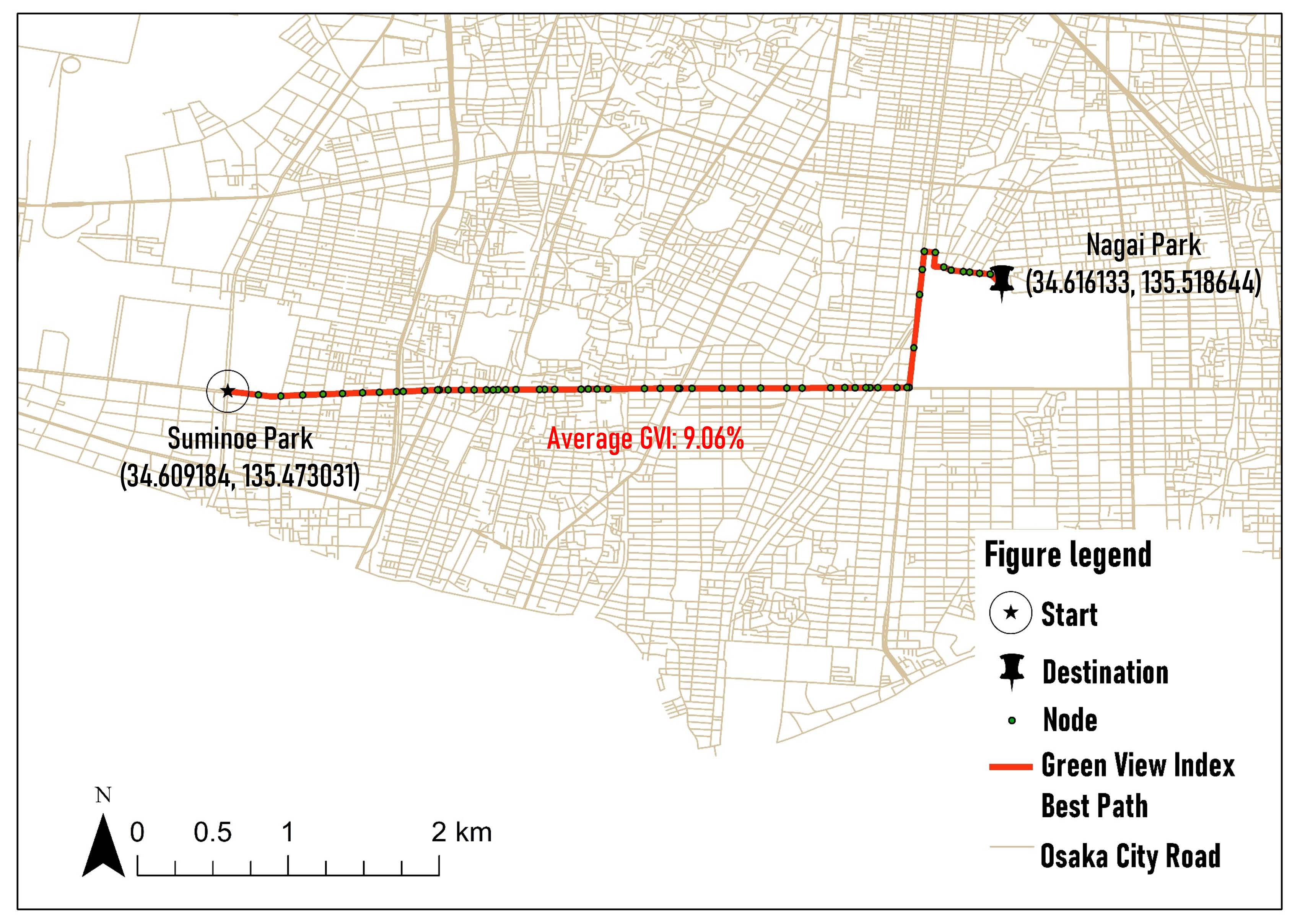}}
\hspace{4mm}
\subfloat[The GVI best path from Nagai Park (34.616133, 135.518644) to Suminoe Park (34.609184, 135.473031), which is the same GVI path as \protect\subref{map1.sub.3}.]{
\label{map1.sub.4}
\includegraphics[width=0.46\linewidth]{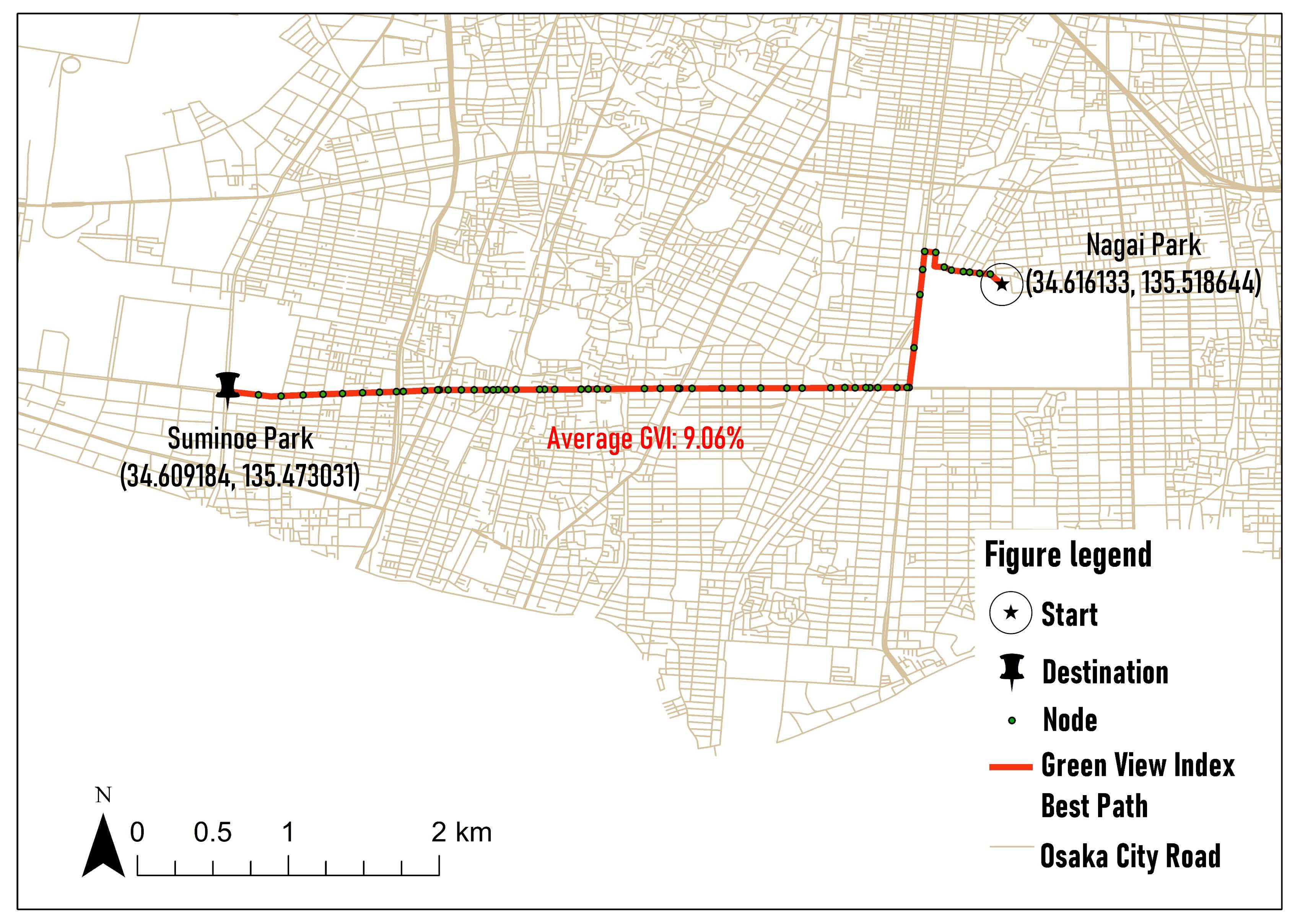}}
\vspace{2mm}
\subfloat[The GVI best path from Tsurumi Ryokuchi Park (34.709121, 135.574583) to Osaka Castle (34.690176, 135.534549). The average GVI of this path is 12.74\%.]{
\label{map1.sub.5}
\includegraphics[width=0.46\linewidth]{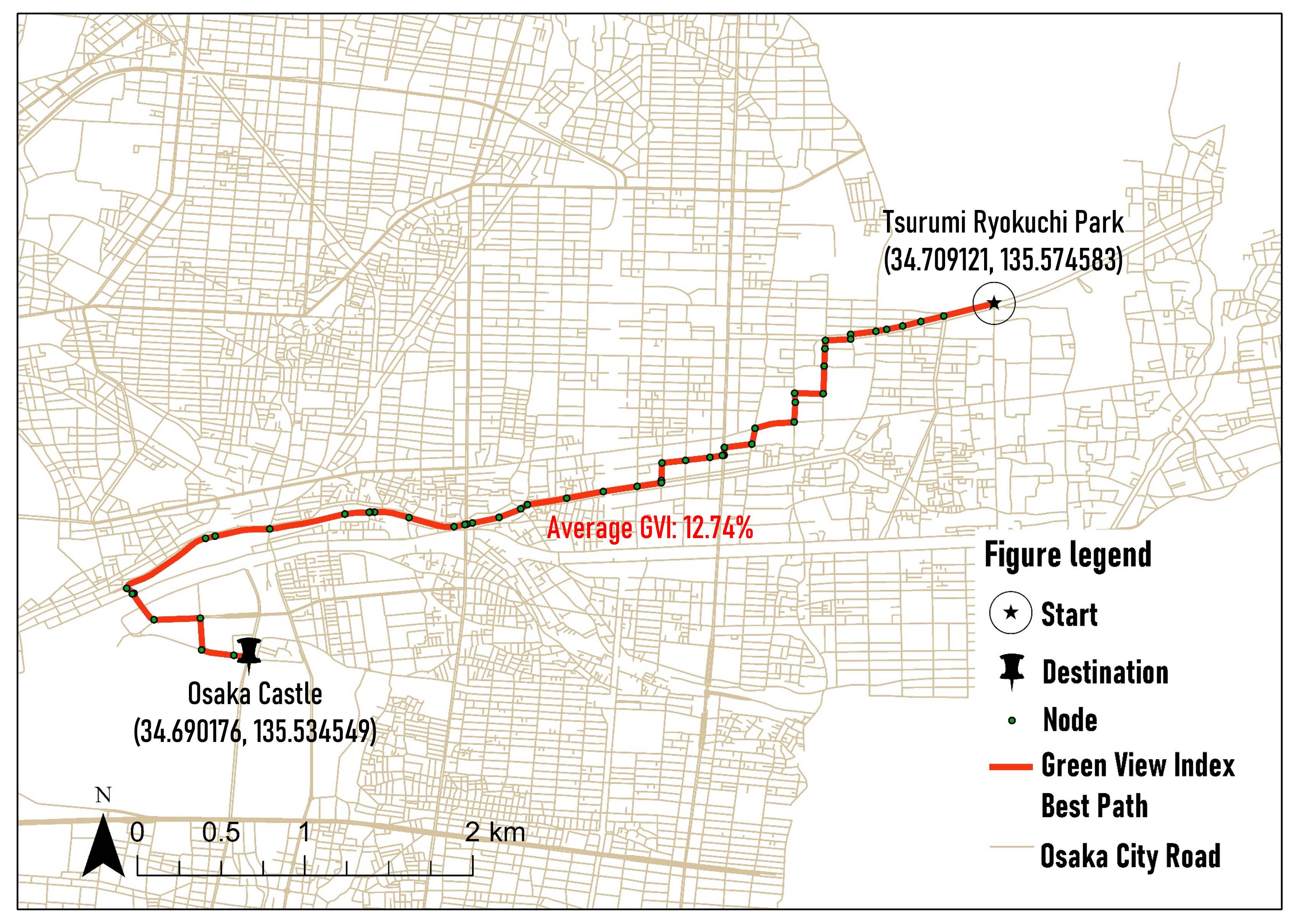}}
\hspace{4mm}
\subfloat[The GVI best path from Osaka Castle (34.690176, 135.534549) to Tsurumi Ryokuchi Park (34.709121, 135.574583). The same GVI path as \protect\subref{map1.sub.5}.]{
\label{map1.sub.6}
\includegraphics[width=0.46\linewidth]{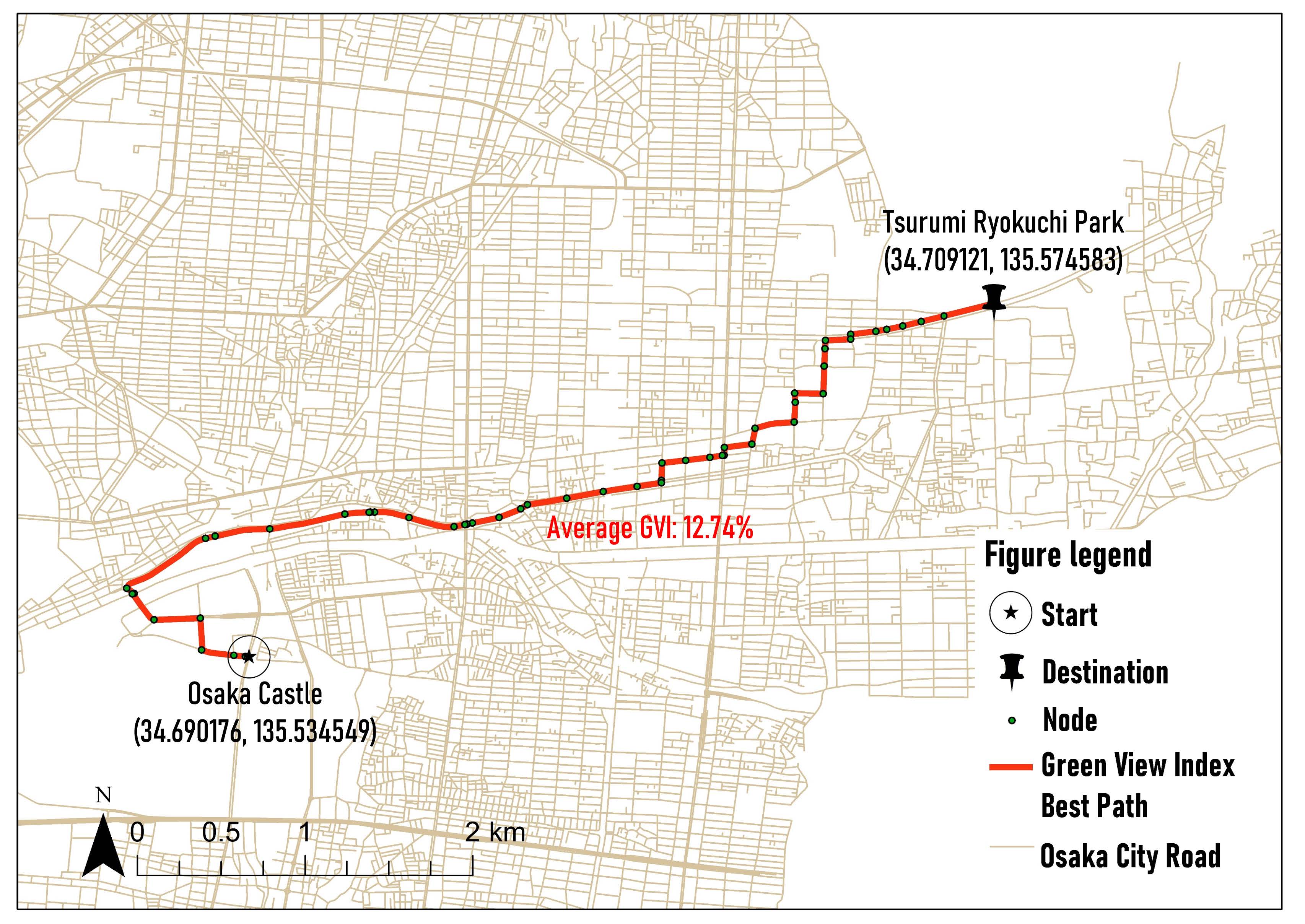}}
\vspace{0.15in}
\caption{Three more experimental best GVI paths in Osaka City with examples of switched starting points and destinations.}
\label{map1}
\end{figure*}

    \subsubsection{Evaluation method of the GVI best paths}
    We use the number of line segments contained in the path and the total GVI of the path as materials to calculate the average GVI of this path as an evaluation method.

    \subsubsection{The examples of the best GVI paths by our approach}
    We choose Nagai Park as the starting point, drive north, and select Utsubo Park as the destination. The obtained GVI best path by the proposed method, as well as the example images of the start, destination, and part of the turning points, are shown in Fig. \ref{fig:Floyd}. Moreover, we show more experiments of the common approach in Fig. \ref{map1}. We switched the selected starting points and destinations to visually demonstrate the correctness of the paths, such as whether they are the same paths and have the same average GVIs. Three paths are included: Namba $\Leftrightarrow$ Umeda (Fig. \ref{map1.sub.1},  Fig. \ref{map1.sub.2}), Suminoe Park $\Leftrightarrow$ Nagai Park (Fig. \ref{map1.sub.3}, Fig. \ref{map1.sub.4}) and Tsurumi Ryokuchi Park $\Leftrightarrow$ Osaka Castle (Fig. \ref{map1.sub.5}, Fig. \ref{map1.sub.6}). 
    
    \subsubsection{Advantages and disadvantages}
    Solving the best GVI path using the adjacency matrix and the Floyd-Warshall Algorithm has the following advantages.
\begin{enumerate}
\item We break through the limitation of ArcGIS software and use the point-line relationship derived from ArcGIS software as the material to solve the practical best GVI path.
\item Floyd-Warshall Algorithm is simple in principle. Once the \textbf{parents} matrix containing the point-line relationships of GVI is solved, the best GVI path is obtained by specifying any starting points and destinations contained in the matrix.
\end{enumerate}

    The disadvantage of this common approach is listed below:
\begin{enumerate}
\item Calculating the \textbf{parents} matrix is very time-consuming. However, once we have the \textbf{parents} matrix for the whole region, we can arbitrarily specify the starting points and destinations to get the best GVI path.
\end{enumerate}

\begin{figure}[bt!]
\centering
\includegraphics[width=1\linewidth]{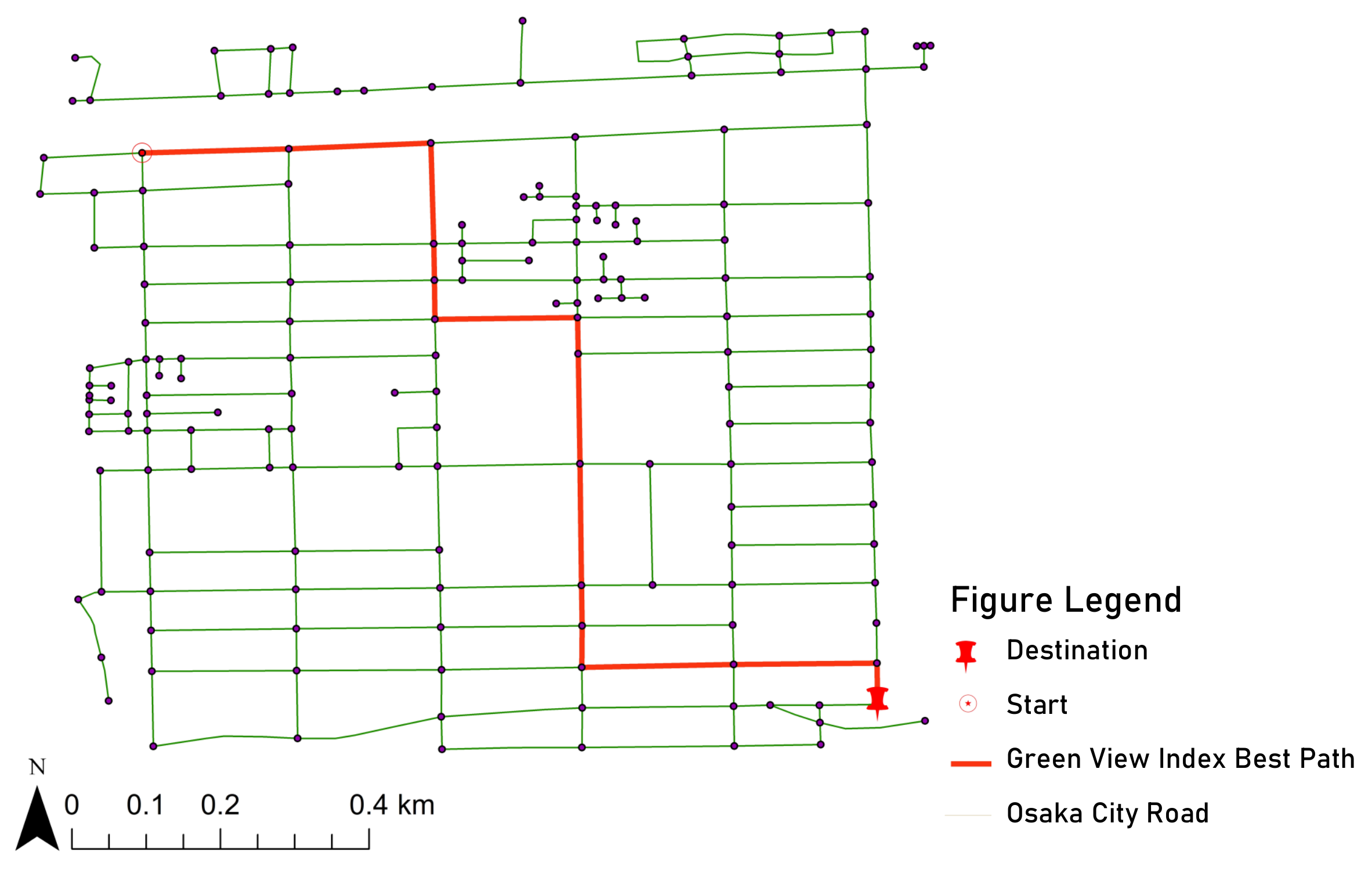}
\vspace{-0.05in}
\caption{Calculation of the GVI best path based on the GVI facing the street direction.}
\label{fig:imazu_path}
\end{figure}

\subsection{A More Realistic Attempt -- With Limitation}
    \subsubsection{Landscape in the direction of travel}
    The GVI${_{avg}}$ of each line is used as a measure to calculate the GVI best path for any location. However, in a continuous landscape of real life, the human eye mainly observes the landscape in the direction of travel, and the average GVI of two nodes is not used to represent the GVI of the line. Therefore, a more accurate method of calculating the GVI best path should be proposed by choosing the direction of each node toward the road to calculate the GVI. The Imazu Park is an ideal area to try this idea because it is almost a square area, while each node towards east, west, north, and south horizontally and vertically, as well as the street view image we intercepted, can be a well-represented view of the people's direction of travel. 
    
    \subsubsection{An ideal place -- Imazu Park}
    In the example of the Imazu Park in Osaka City, we collected street view images from the road angle and processed them to obtain the GVI of each intersection direction, following the method of collecting and analyzing street view images in Osaka City. We obtained 0$^\circ$, 90$^\circ$, 180$^\circ$, and 270$^\circ$ street view images of each intersection. Concerning the treatment of the adjacency matrix, we adopted a bi-directional directed graph instead of the previous undirected graph to include more information on weights. The best path was calculated based on the street direction GVI as shown in Fig. \ref{fig:imazu_path}.
    
    \subsubsection{Pros and Cons}
    This method best matches the realistic scene of the travel route, i.e., the GVI is different between two nodes with different orientations. However, this method has a significant limitation in that we cannot directly get the street view image toward the direction of the path and calculate the GVI path due to the limitation of Google API. We only tried on a small square area and tried to generalize this idea.

\begin{figure}[bt!]
\centering
\includegraphics[width=1\linewidth]{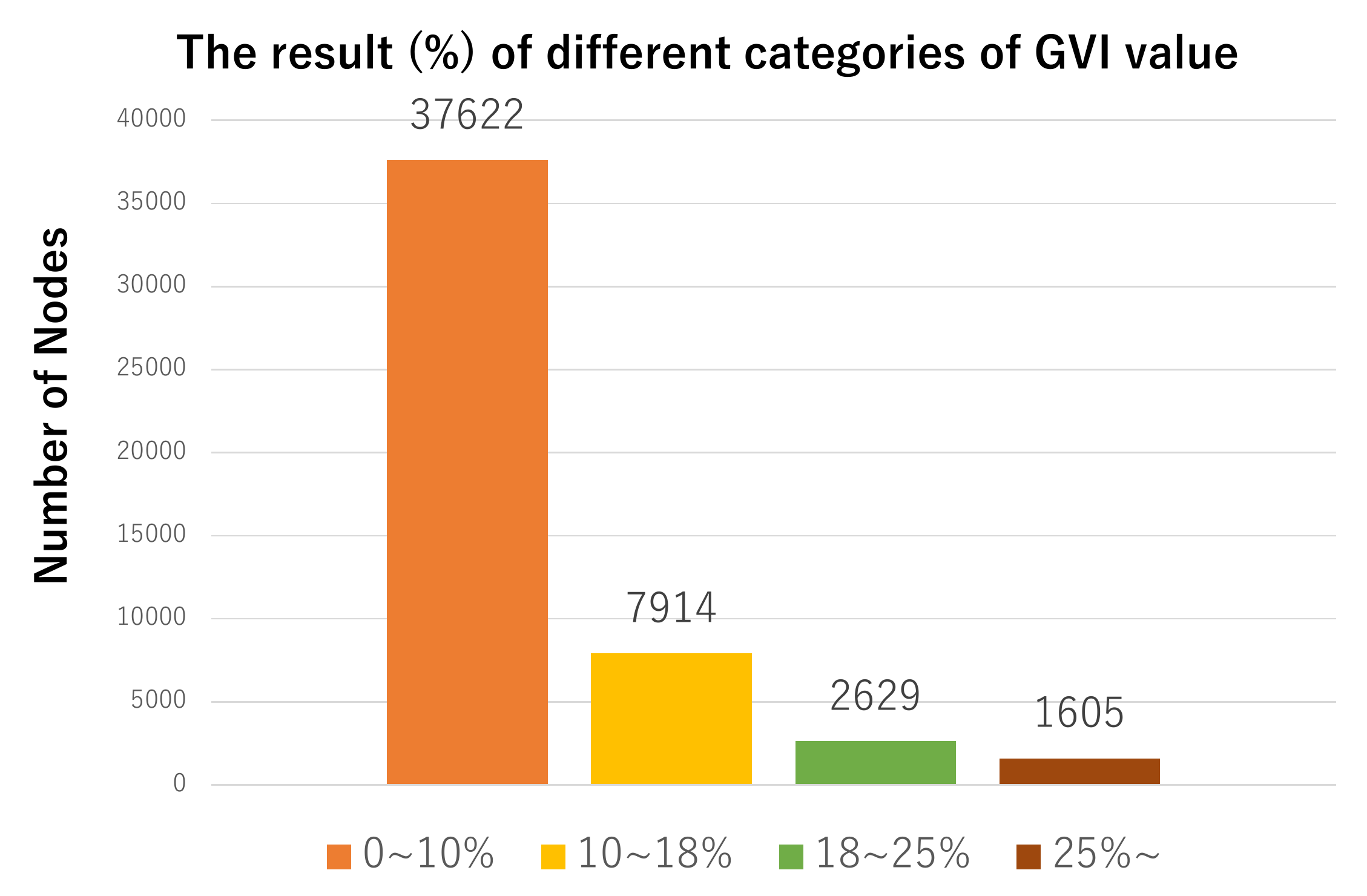}
\caption{The result of different categories of GVI in Osaka City.}
\label{fig:satis}
\end{figure}

\section{Discussion and Result}
\subsection{The GVI Analysis of Osaka City}
\begin{figure}[bt!]
\centering
\includegraphics[width=1\linewidth]{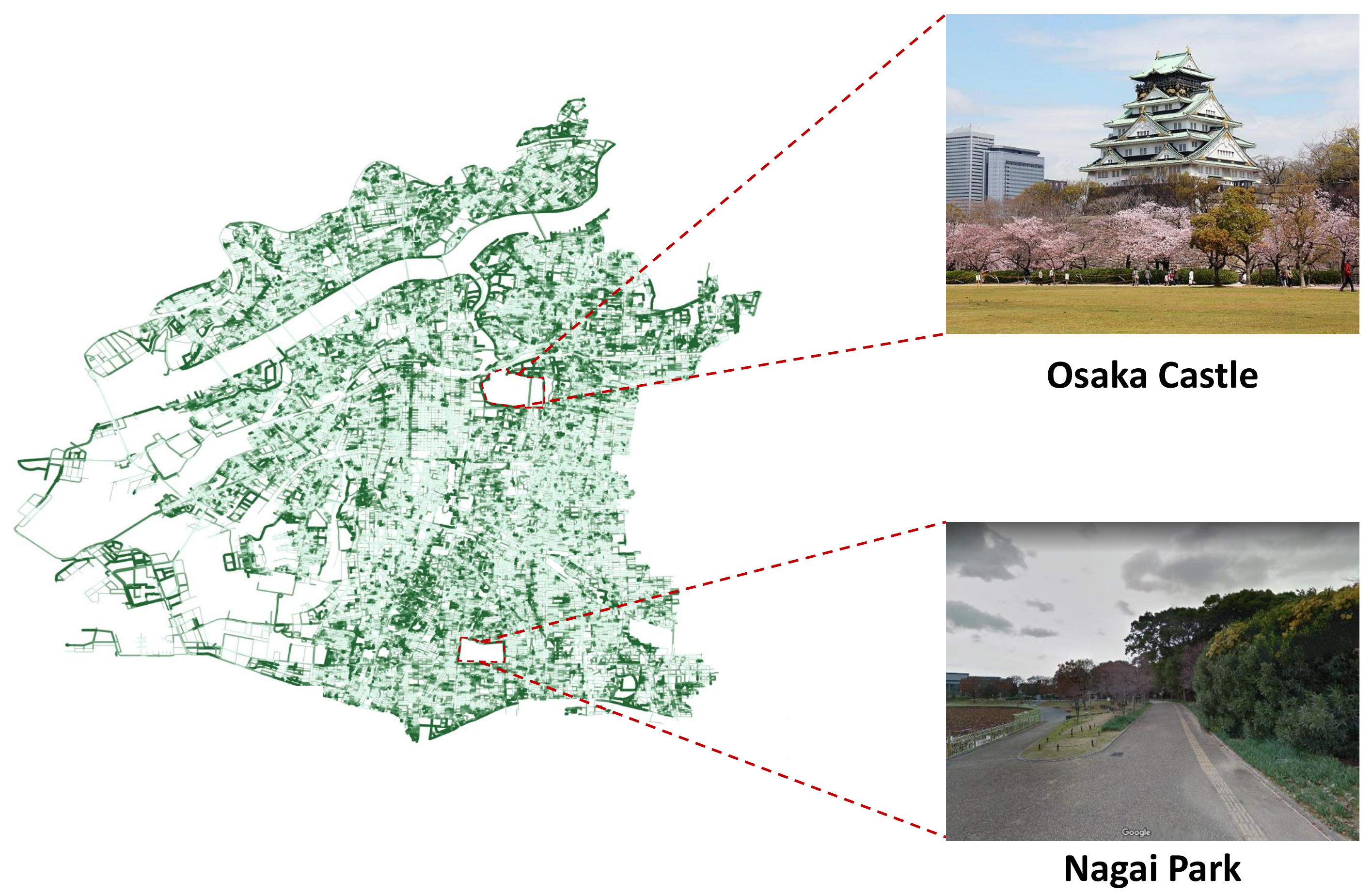}
\vspace{0.05in}
\caption{Areas not captured in the GVI distribution map: Osaka Castle and Nagai Park.}
\label{fig:no_green}
\end{figure}

    In the part of GVI analysis in this paper, as shown in Fig. \ref{fig:satis}, there are 37,622 nodes with an average GVI between 0$\sim$10\% of the 49,770 nodes in Osaka City overall, accounting for 75.59\% of the total; there are 7,914 nodes with an average GVI between 10$\sim$18\%, representing 15.9\% of the total; 2,629 nodes with an average GVI between 18$\sim$25\%, which represent 5.28\% of the total; 1,605 nodes (3.22\% of the total) have an average GVI of 25\%$\sim$. From the result of statistics based on HRNet-OCR, Osaka City has a low level of GVI and Low satisfaction based on the percentage of GVI over 25\%. Since we used the "drive" mode to collect geographic data, we did not capture available street view images and calculate the corresponding GVI for the regions that were not passable by vehicles but were passable by pedestrians on foot. The most prominent areas, such as Osaka Castle and Nagai Park, are not displayed on the GVI distribution map (see Fig. \ref{fig:no_green}).
    
    Based on the distribution data of the GVI in Osaka City, the results of calculating the GVI best path can be derived. The following summarizes the results of three different methods: GVI best path by ArcGIS, adjacency matrix, and Floyd-Warshall Algorithm, as well as the more realistic attempt.

\subsection{The GVI Best Path of Osaka City}
    The threshold output produced in the reference corridor analysis can be considered a least-cost corridor for image elements rather than a least-cost path. From Nagai Park to Utsubo Park, it is the path zone, not the specific route, which is planned. We can only obtain a general path zone, which does not apply to actual tourist routes due to the lack of sufficient detail and the nature of travel along the path. However, we are inspired by the fact that this path zone allows us to roughly derive the direction of travel of the best GVI path, and it is a direction that is worth considering and practicing whether this approximate direction of travel can be applied to optimal GVI path analysis when operations such as the path adjacency matrix are too computationally intensive. For the second method, the geometric network analysis method, taking the path from Nagai Park to Utsubo Park as an example, the geometric network analysis method applies to designing multiple stopping nodes. However, this method is not feasible as we cannot obtain the correct GVI path due to the lack of vital information on the number of nodes through which the path passes. Nevertheless, this method, known as network analysis, can be used not only for planning tourist routes in the area but also for evaluating neighborhood landscapes and the evaluation of business values in the area.

    We used the point-line relationships derived from ArcGIS software and generated the corresponding adjacency matrix. Floyd-Warshall Algorithm is a feasible method, and we finally get the correct GVI best path. This method can be used in urban tourism route design, marathon race route design, and other situations where the urban landscape needs to be displayed. However, we are considering other methods to save computation time and improve efficiency due to its long computation time. We have verified the correctness of the GVI best path, and the most intuitive way to gain an intuitive understanding is by comparing the GVI of neighboring roads with it. The adjacency matrix allows the coordinates and GVI of the nodes through which the path passes to be returned. As shown in Fig. \ref{fig:Floyd}, we randomly selected street view images of several inflection nodes and aligned them. This best GVI path passes through a total of 120 nodes (including starting point and destination), and the average GVI of this path was 7.47\%. We also conducted more experiments to prove the generalizability of our proposed method, as shown in Fig. \ref{map1}. We randomly selected attractions and commercial centers located in different locations as starting points and destinations to calculate the GVI best paths. Furthermore, we switched them to facilitate a more visual presentation of the results of the GVI best path.

    For the realistic assignment method, the GVI best path is calculated based on the directional GVI. Take the example of the northwest node to the southeast node of Imazu Park in Osaka City. Using ArcGIS software, we can obtain the GVI best path using the traditional method of converting the best GVI path problem into a shortest path problem. In this area, because of its square shape and the coincidental direction of the roads (basically, the roads are facing due north, south, west, and east.), there is no limit to the number of nodes so that the correct best GVI path can be obtained with conventional ArcGIS. We have studied this area using a generic adjacency matrix and the Floyd-Warshall Algorithm. The GVI assignment of the roads was performed using a different strategy, i.e., a street view image of each node facing the street was calculated as a GVI assignment to the lines instead of the previous assignment of each line by the average of the nodes connected to it. Because the detailed GVI is calculated for the different directions of each node, we can obtain a more realistic GVI best path, as in a real-world implementation. However, this method is quite cumbersome in GVI statistics. As the roads in Imazu Park intersect each other vertically, it is possible to set up the download of street view images at 0$^\circ$, 90$^\circ$, 180$^\circ$, and 270$^\circ$. Regarding the part of the adjacency matrix, instead of undirected maps, we use directed maps with different directions to generate the adjacency matrix so that there are two directions with different weights for each line, and accordingly, the cost of the calculation will be increased. This approach to achieving the best GVI path to match realistic scenes is not easy to implement because it has many strict constraints, such as the requirement to find the angle of the road direction for downloading street view images, and the adjacency matrix needs to be set as a directed graph, leading to an increase in the cost of computing the path. This method could be better improved and generalized if an optional API exists in GSV for the angle of the road to the coordinate system for acquiring street view images. Otherwise, it is worth exploring in our future research to use two nodes close to each other to calculate the angle with the coordinate system to find the road orientation. Besides, to fill in the gaps of non-drivable roads, we will collect data resources in a walkable way in the future to get more complete results.

\section{Conclusion}
    This paper implements a semantic segmentation network in deep learning to analyze the Green View Index in Osaka City. We implement the GVI analysis on a large scale, get satisfactory results and visualize them with feasible methods. After obtaining the preliminary GVI distribution of Osaka city, we proposed an innovative generic approach to calculate the GVI best path and verified it more carefully and accurately in a limited area. Not only did we implement the GVI best path by street view images, filling a gap in this application, but we also demonstrated the feasibility of this research, which allows us to guide better lifestyles through street view images.

\section*{Acknowledgements}
This work was supported by Ms. Jinsui, Zhao, and Mr. Jifang, Zhang. Thanks to the reviewers for their constructive suggestions.

\section*{Conflict of interest statement}
All the authors declare that there is no actual or potential conflict of interest, including any financial, personal, or other relationships with other people or organizations.

{\small
\bibliographystyle{unsrt}
\bibliography{main_paper}

\begin{thebibliography}{10}

\bibitem{li2015assessing}
Xiaojiang Li, Chuanrong Zhang, Weidong Li, Robert Ricard, Qingyan Meng, and
  Weixing Zhang.
\newblock Assessing street-level urban greenery using google street view and a
  modified green view index.
\newblock {\em Urban Forestry \& Urban Greening}, 14(3):675--685, 2015.

\bibitem{1de2013streetscape}
Sjerp De~Vries, Sonja~Me Van~Dillen, Peter~P Groenewegen, and Peter
  Spreeuwenberg.
\newblock Streetscape greenery and health: Stress, social cohesion and physical
  activity as mediators.
\newblock {\em Social science \& medicine}, 94:26--33, 2013.

\bibitem{2mcpherson2011million}
E~Gregory McPherson, James~R Simpson, Qingfu Xiao, and Chunxia Wu.
\newblock Million trees los angeles canopy cover and benefit assessment.
\newblock {\em Landscape and Urban Planning}, 99(1):40--50, 2011.

\bibitem{3van2016view}
Timothy Van~Renterghem and Dick Botteldooren.
\newblock View on outdoor vegetation reduces noise annoyance for dwellers near
  busy roads.
\newblock {\em Landscape and urban planning}, 148:203--215, 2016.

\bibitem{4ferrini2020role}
Francesco Ferrini, Alessio Fini, Jacopo Mori, and Antonella Gori.
\newblock Role of vegetation as a mitigating factor in the urban context.
\newblock {\em Sustainability}, 12(10):4247, 2020.

\bibitem{11liu2017machine}
Lun Liu, Elisabete~A Silva, Chunyang Wu, and Hui Wang.
\newblock A machine learning-based method for the large-scale evaluation of the
  qualities of the urban environment.
\newblock {\em Computers, environment and urban systems}, 65:113--125, 2017.

\bibitem{6zhang2014cooling}
Biao Zhang, Ji-xi Gao, Yang Yang, et~al.
\newblock The cooling effect of urban green spaces as a contribution to
  energy-saving and emission-reduction: A case study in beijing, china.
\newblock {\em Building and environment}, 76:37--43, 2014.

\bibitem{runGVI}
Dengkai Huang, Bin Jiang, and Lei Yuan.
\newblock Analyzing the effects of nature exposure on perceived satisfaction
  with running routes: An activity path-based measure approach.
\newblock {\em Urban Forestry \& Urban Greening}, page 127480, 2022.

\bibitem{remoteimage2019}
Henri Riihim{\"a}ki, Miska Luoto, and Janne Heiskanen.
\newblock Estimating fractional cover of tundra vegetation at multiple scales
  using unmanned aerial systems and optical satellite data.
\newblock {\em Remote Sensing of Environment}, 224:119--132, 2019.

\bibitem{rsistudy2020}
Elena Barbierato, Iacopo Bernetti, Irene Capecchi, and Claudio Saragosa.
\newblock Integrating remote sensing and street view images to quantify urban
  forest ecosystem services.
\newblock {\em Remote sensing}, 12(2):329, 2020.

\bibitem{streetviewgis2021}
Filip Biljecki and Koichi Ito.
\newblock Street view imagery in urban analytics and gis: A review.
\newblock {\em Landscape and Urban Planning}, 215:104217, 2021.

\bibitem{dong2018green}
Rencai Dong, Yonglin Zhang, and Jingzhu Zhao.
\newblock How green are the streets within the sixth ring road of beijing? an
  analysis based on tencent street view pictures and the green view index.
\newblock {\em International journal of environmental research and public
  health}, 15(7):1367, 2018.

\bibitem{8yang2009can}
Jun Yang, Linsen Zhao, Joe Mcbride, and Peng Gong.
\newblock Can you see green? assessing the visibility of urban forests in
  cities.
\newblock {\em Landscape and Urban Planning}, 91(2):97--104, 2009.

\bibitem{ye2019urban}
Nanqi Ye, Bowen Wang, Michihiro Kita, Ming Xie, and Wenyue Cai.
\newblock Urban commerce distribution analysis based on street view and deep
  learning.
\newblock {\em IEEE Access}, 7:162841--162849, 2019.

\bibitem{wang2021noisy}
Bowen Wang, Liangzhi Li, Yuta Nakashima, Ryo Kawasaki, Hajime Nagahara, and
  Yasushi Yagi.
\newblock Noisy-lstm: Improving temporal awareness for video semantic
  segmentation.
\newblock {\em IEEE Access}, 9:46810--46820, 2021.

\bibitem{7tong2020evaluating}
Ming Tong, Jiangfeng She, Junzhong Tan, Mengyao Li, Rongcun Ge, and Yiyuan Gao.
\newblock Evaluating street greenery by multiple indicators using street-level
  imagery and satellite images: A case study in nanjing, china.
\newblock {\em Forests}, 11(12):1347, 2020.

\bibitem{longestpth}
Ryuhei Uehara and Yushi Uno.
\newblock Efficient algorithms for the longest path problem.
\newblock In {\em International symposium on algorithms and computation}, pages
  871--883. Springer, 2004.

\bibitem{shortestalgo}
Kairanbay Magzhan and Hajar~Mat Jani.
\newblock A review and evaluations of shortest path algorithms.
\newblock {\em International journal of scientific \& technology research},
  2(6):99--104, 2013.

\bibitem{HRNet-OCR}
Yuhui Yuan, Xiaokang Chen, Xilin Chen, and Jingdong Wang.
\newblock Segmentation transformer: Object-contextual representations for
  semantic segmentation.
\newblock {\em arXiv preprint arXiv:1909.11065}, 2019.

\bibitem{zjxmdpi}
Jiaxin Zhang, Tomohiro Fukuda, and Nobuyoshi Yabuki.
\newblock Development of a city-scale approach for fa{\c{c}}ade color
  measurement with building functional classification using deep learning and
  street view images.
\newblock {\em ISPRS International Journal of Geo-Information}, 10(8):551,
  2021.

\bibitem{lyq2022}
Yunqin Li, Nobuyoshi Yabuki, and Tomohiro Fukuda.
\newblock Exploring the association between street built environment and street
  vitality using deep learning methods.
\newblock {\em Sustainable Cities and Society}, 79:103656, 2022.

\bibitem{zjxieee}
Jiaxin Zhang, Tomohiro Fukuda, and Nobuyoshi Yabuki.
\newblock Automatic object removal with obstructed fa{\c{c}}ades completion
  using semantic segmentation and generative adversarial inpainting.
\newblock {\em IEEE Access}, 9:117486--117495, 2021.

\bibitem{16kang2018building}
Jian Kang, Marco K{\"o}rner, Yuanyuan Wang, Hannes Taubenb{\"o}ck, and
  Xiao~Xiang Zhu.
\newblock Building instance classification using street view images.
\newblock {\em ISPRS journal of photogrammetry and remote sensing}, 145:44--59,
  2018.

\bibitem{2.1.1}
Iana Markevych, Julia Schoierer, Terry Hartig, Alexandra Chudnovsky, Perry
  Hystad, Angel~M Dzhambov, Sjerp De~Vries, Margarita Triguero-Mas, Michael
  Brauer, Mark~J Nieuwenhuijsen, et~al.
\newblock Exploring pathways linking greenspace to health: Theoretical and
  methodological guidance.
\newblock {\em Environmental research}, 158:301--317, 2017.

\bibitem{2.1.2}
I-Min Lee, Eric~J Shiroma, Felipe Lobelo, Pekka Puska, Steven~N Blair, Peter~T
  Katzmarzyk, Lancet Physical Activity Series~Working Group, et~al.
\newblock Effect of physical inactivity on major non-communicable diseases
  worldwide: an analysis of burden of disease and life expectancy.
\newblock {\em The lancet}, 380(9838):219--229, 2012.

\bibitem{gviphsical}
Yi~Lu.
\newblock Using google street view to investigate the association between
  street greenery and physical activity.
\newblock {\em Landscape and Urban Planning}, 191:103435, 2019.

\bibitem{walk}
Donghwan Ki and Sugie Lee.
\newblock Analyzing the effects of green view index of neighborhood streets on
  walking time using google street view and deep learning.
\newblock {\em Landscape and Urban Planning}, 205:103920, 2021.

\bibitem{DNN}
Alfredo Canziani, Adam Paszke, and Eugenio Culurciello.
\newblock An analysis of deep neural network models for practical applications.
\newblock {\em arXiv preprint arXiv:1605.07678}, 2016.

\bibitem{ma2022building}
Kai Ma, Bowen Wang, Yunqin Li, and Jiaxin Zhang.
\newblock Image retrieval for local architectural heritage recommendation based
  on deep hashing.
\newblock {\em Buildings}, 12(6):809, 2022.

\bibitem{23zhao2017pyramid}
Hengshuang Zhao, Jianping Shi, Xiaojuan Qi, Xiaogang Wang, and Jiaya Jia.
\newblock Pyramid scene parsing network.
\newblock In {\em Proceedings of the IEEE conference on computer vision and
  pattern recognition}, pages 2881--2890, 2017.

\bibitem{17zhang2019concrete}
Xinxiang Zhang, Dinesh Rajan, and Brett Story.
\newblock Concrete crack detection using context-aware deep semantic
  segmentation network.
\newblock {\em Computer-Aided Civil and Infrastructure Engineering},
  34(11):951--971, 2019.

\bibitem{wang2021automatic}
Bowen Wang, Toshihiro Takeda, Kento Sugimoto, Jiahao Zhang, Shoya Wada, Shozo
  Konishi, Shirou Manabe, Katsuki Okada, and Yasushi Matsumura.
\newblock Automatic creation of annotations for chest radiographs based on the
  positional information extracted from radiographic image reports.
\newblock {\em Computer Methods and Programs in Biomedicine}, 209:106331, 2021.

\bibitem{19siam2017deep}
Mennatullah Siam, Sara Elkerdawy, Martin Jagersand, and Senthil Yogamani.
\newblock Deep semantic segmentation for automated driving: Taxonomy, roadmap
  and challenges.
\newblock In {\em 2017 IEEE 20th international conference on intelligent
  transportation systems (ITSC)}, pages 1--8. IEEE, 2017.

\bibitem{20ronneberger2015u}
Olaf Ronneberger, Philipp Fischer, and Thomas Brox.
\newblock U-net: Convolutional networks for biomedical image segmentation.
\newblock In {\em International Conference on Medical image computing and
  computer-assisted intervention}, pages 234--241. Springer, 2015.

\bibitem{21badrinarayanan2017segnet}
Vijay Badrinarayanan, Alex Kendall, and Roberto Cipolla.
\newblock Segnet: A deep convolutional encoder-decoder architecture for image
  segmentation.
\newblock {\em IEEE transactions on pattern analysis and machine intelligence},
  39(12):2481--2495, 2017.

\bibitem{deeplabv3+}
Liang-Chieh Chen, Yukun Zhu, George Papandreou, Florian Schroff, and Hartwig
  Adam.
\newblock Encoder-decoder with atrous separable convolution for semantic image
  segmentation.
\newblock In {\em Proceedings of the European conference on computer vision
  (ECCV)}, pages 801--818, 2018.

\bibitem{cordts2016cityscapes}
Marius Cordts, Mohamed Omran, Sebastian Ramos, Timo Rehfeld, Markus Enzweiler,
  Rodrigo Benenson, Uwe Franke, Stefan Roth, and Bernt Schiele.
\newblock The cityscapes dataset for semantic urban scene understanding.
\newblock In {\em Proceedings of the IEEE conference on computer vision and
  pattern recognition}, pages 3213--3223, 2016.

\bibitem{HRNet}
Ke~Sun, Bin Xiao, Dong Liu, and Jingdong Wang.
\newblock Deep high-resolution representation learning for human pose
  estimation.
\newblock In {\em CVPR}, 2019.

\bibitem{newyork}
Xiaojiang Li.
\newblock Examining the spatial distribution and temporal change of the green
  view index in new york city using google street view images and deep
  learning.
\newblock {\em Environment and Planning B: Urban Analytics and City Science},
  48(7):2039--2054, 2021.

\bibitem{yokohama}
Yusuke Kumakoshi, Sau~Yee Chan, Hideki Koizumi, Xiaojiang Li, and Yuji
  Yoshimura.
\newblock Standardized green view index and quantification of different metrics
  of urban green vegetation.
\newblock {\em Sustainability}, 12(18):7434, 2020.

\bibitem{25cetin2015using}
Mehmet Cetin.
\newblock Using gis analysis to assess urban green space in terms of
  accessibility: case study in kutahya.
\newblock {\em International Journal of Sustainable Development \& World
  Ecology}, 22(5):420--424, 2015.

\bibitem{26yamagata2016value}
Yoshiki Yamagata, Daisuke Murakami, Takahiro Yoshida, Hajime Seya, and Sho
  Kuroda.
\newblock Value of urban views in a bay city: Hedonic analysis with the spatial
  multilevel additive regression (smar) model.
\newblock {\em Landscape and Urban Planning}, 151:89--102, 2016.

\bibitem{greenpath}
Joose Helle, Age Poom, Elias~S Willberg, and Tuuli Toivonen.
\newblock The green paths route planning software for exposure-optimised
  travel.
\newblock 2021.

\bibitem{omsgreen}
Tessio Novack, Zhiyong Wang, and Alexander Zipf.
\newblock A system for generating customized pleasant pedestrian routes based
  on openstreetmap data.
\newblock {\em Sensors}, 18(11):3794, 2018.

\bibitem{2.3.1}
Paulo Ribeiro and Jos{\'e}~FG Mendes.
\newblock Route planning for soft modes of transport: healthy routes.
\newblock {\em WIT Transactions on the Built Environment}, 116:677--688, 2011.

\bibitem{heatstress}
Joachim Ru{\ss}ig and Julian Bruns.
\newblock Reducing individual heat stress through path planning.
\newblock {\em GI\_Forum}, 1:327--340, 2017.

\bibitem{hernandez2018allergyless}
Mar{\'{i}}a~Luisa Hern{\'a}ndez-Alcaraz, Isabel~Mar{\'{i}}a Robles-Mar{\'{i}}n,
  Francisco Garc{\'{i}}a-S{\'a}nchez, and Rafael Valencia-Garc{\'{i}}a.
\newblock Allergyless. an intelligent recommender system to reduce exposition
  time to allergens in smart-cities.
\newblock In {\em Distributed Computing and Artificial Intelligence, 15th
  International Conference}, volume 800, page~61. Springer, 2018.

\bibitem{28boeing2017osmnx}
Geoff Boeing.
\newblock Osmnx: New methods for acquiring, constructing, analyzing, and
  visualizing complex street networks.
\newblock {\em Computers, Environment and Urban Systems}, 65:126--139, 2017.

\bibitem{OCRNet}
Yuhui Yuan, Xilin Chen, and Jingdong Wang.
\newblock Object-contextual representations for semantic segmentation.
\newblock In {\em European conference on computer vision}, pages 173--190.
  Springer, 2020.

\end{thebibliography}
}

\end{document}